\RequirePackage{fix-cm}
\documentclass[twocolumn]{svjour3}          % twocolumn
\smartqed  % flush right qed marks, e.g. at end of proof
\usepackage{graphicx}
\usepackage{ifsym}

\usepackage{amssymb}
\usepackage{amsmath}
\usepackage{mathtools} 
\usepackage[margin=0.75 in]{geometry}
\usepackage[utf8]{inputenc}
\usepackage{times}  % DO NOT CHANGE THIS
\usepackage{helvet} % DO NOT CHANGE THIS
\usepackage{courier}  % DO NOT CHANGE THIS
\usepackage[hyphens]{url}  % DO NOT CHANGE THIS
\usepackage{graphicx} % DO NOT CHANGE THIS
\usepackage[sectionbib,square]{natbib}  % DO NOT CHANGE THIS AND DO NOT ADD ANY OPTIONS TO IT
\usepackage{caption} % DO NOT CHANGE THIS AND DO NOT ADD ANY OPTIONS TO IT
\usepackage{subfigure}
\usepackage{booktabs}
\usepackage{xcolor}
\usepackage{tabularx}
\usepackage{multirow}
\usepackage{hyperref}
\usepackage{lipsum}

\usepackage{amssymb}
\usepackage{array}

\begin{document}

\title{Spoofing Detection on Hand Images Using Quality Assessment
}
\author{Asish Bera \textsuperscript{*} 
\and
        Ratnadeep Dey \textsuperscript{*} 
        \and 
        Debotosh Bhattacharjee %\textsuperscript{2,3}
        \and 
        \\
         Mita Nasipuri%\textsuperscript{2}  
         \and 
        Hubert P. H. Shum%\textsuperscript{4}
}

\institute{ A. Bera (Corresponding Author) %\textsuperscript{*}  
\at
              Department of Computer Science, Edge Hill University,  Ormskirk, Lancashire, L39 4QP, U.K. \\
              %Tel.: +123-45-678910\\
              %Fax: +123-45-678910\\
              \email{asish.bera@gmail.com}           
           \and
           R. Dey  %\textsuperscript{*} 
           \at
              Department of Computer Science and Engineering, Jadavpur University, Kolkata-700032, India\\
               \email{ratnadipdey@gmail.com}
               \and
           D. Bhattacharjee  \at
              Department of Computer Science and Engineering, Jadavpur University, Kolkata-700032, India\\ Center for Basic and Applied Science, Faculty of informatics and management, University of Hradec Kralove, Rokitanskeho 62, 500 03 Hradec Kralove, Czech Republic \\
               \email{debotosh@ieee.org}
               \and
           M. Nasipuri\at
              Department of Computer Science and Engineering, Jadavpur University, Kolkata-700032, India\\
               \email{ mitanasipuri@gmail.com}
               \and
           H. P. H. Shum \at
              Department of Computer Science, Durham University, Stockton Road, Durham, DH1 3LE, U.K.\\
               \email{hubert.shum@durham.ac.uk\\\\
               * Both Authors contributed equally to this work.}
}

\date{Received: 30 June 2020 / Accepted: 10 March 2021 / Published Online: 28 May 2021 }
% The correct dates will be entered by the editor

\maketitle
\begin{abstract}
Recent research on biometrics focuses on achieving a high success rate of authentication and addressing the concern of various spoofing attacks. Although hand geometry recognition provides adequate security over unauthorized access, it is susceptible to presentation attack. This paper presents an anti-spoofing method toward hand biometrics. A presentation attack detection approach is addressed by assessing the visual quality of genuine and fake hand images. A threshold-based gradient magnitude similarity quality metric is proposed to discriminate between the real and spoofed hand samples. The visual hand images of 255 subjects from the Bogazici University hand database are considered as original samples. Correspondingly, from each genuine sample, we acquire a forged image using a Canon EOS 700D camera. Such fake hand images with natural degradation are considered for electronic screen display based spoofing attack detection. Furthermore, we create another fake hand dataset with artificial degradation by introducing additional Gaussian blur, salt and pepper, and speckle noises to original images. Ten quality metrics are measured from each sample for classification between original and fake hand image. The classification experiments are performed using the k-nearest neighbors, random forest, and support vector machine classifiers, as well as deep convolutional neural networks. The proposed gradient similarity-based quality metric achieves 1.5\% average classification error using the k-nearest neighbors and random forest  classifiers. An average classification error of 2.5\% is obtained using the baseline evaluation with the MobileNetV2 deep network for discriminating original and different types of fake hand samples.

% \PACS{PACS code1 \and PACS code2 \and more}
% \subclass{MSC code1 \and MSC code2 \and more}
\end{abstract}

\section{Introduction}
\label{intro}
\begin{figure*}
\centering
% Use the relevant command to insert your figure file.
% For example, with the graphicx package use
  \includegraphics[width=0.9\textwidth] {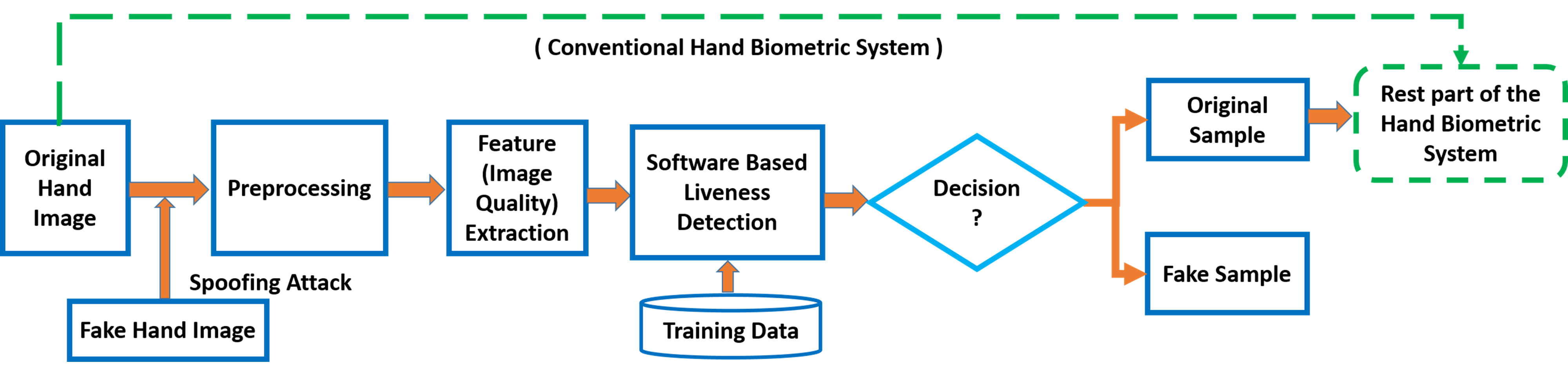}
% figure caption is below the figure
\caption{The proposed anti-spoofing method using hand images. It is pertinent before conventional biometric authentication (specified with a dotted line). A traditional biometric system considers a valid image to discriminate the legitimate and zero-effort imposter, other than the fake sample detection. The feature extraction module is used for conventional hand-crafted feature extraction or deep feature extraction using the MobileNetV2 \textcolor{blue}{\cite{sandler2018mobilenetv2}} base CNN in our proposed scheme.}
\label{fig:main}       % Give a unique label
\end{figure*}

\begin{figure}
\centering
% Use the relevant command to insert your figure file.
% For example, with the graphicx package use
\subfigure []{
  \includegraphics[width=0.40\textwidth, height=3.0 cm] {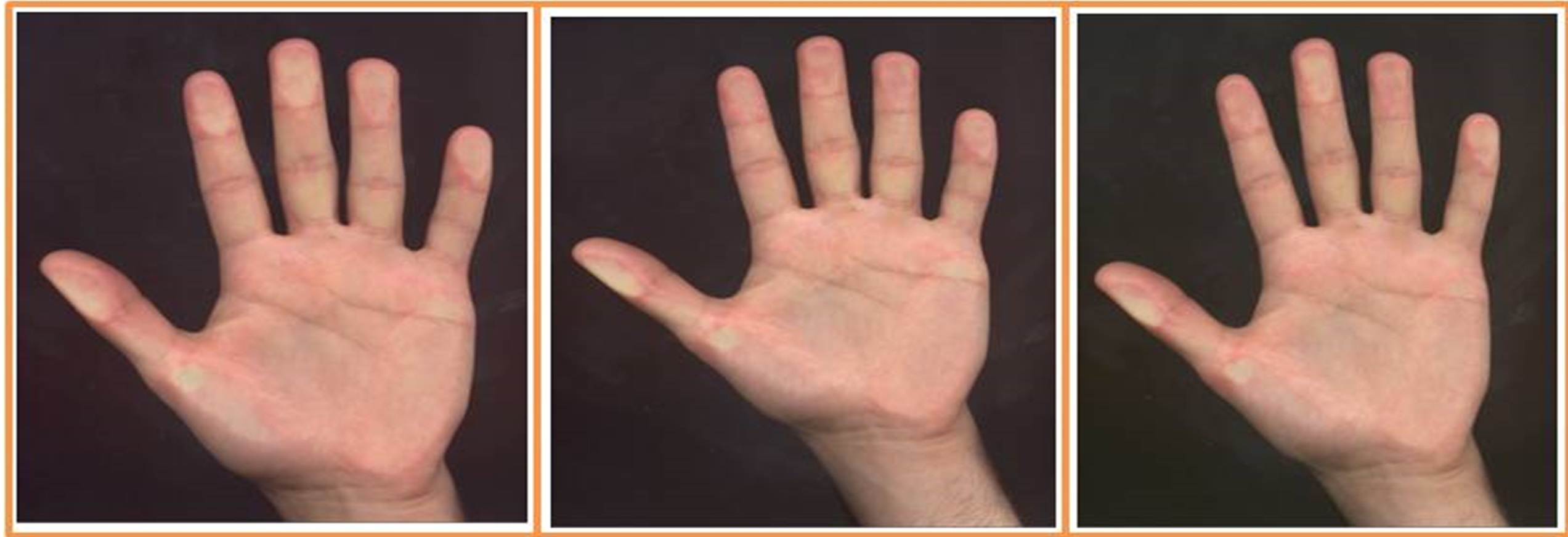}
} 

\subfigure[]{
    \includegraphics[width=0.40\textwidth, height=3.0 cm] {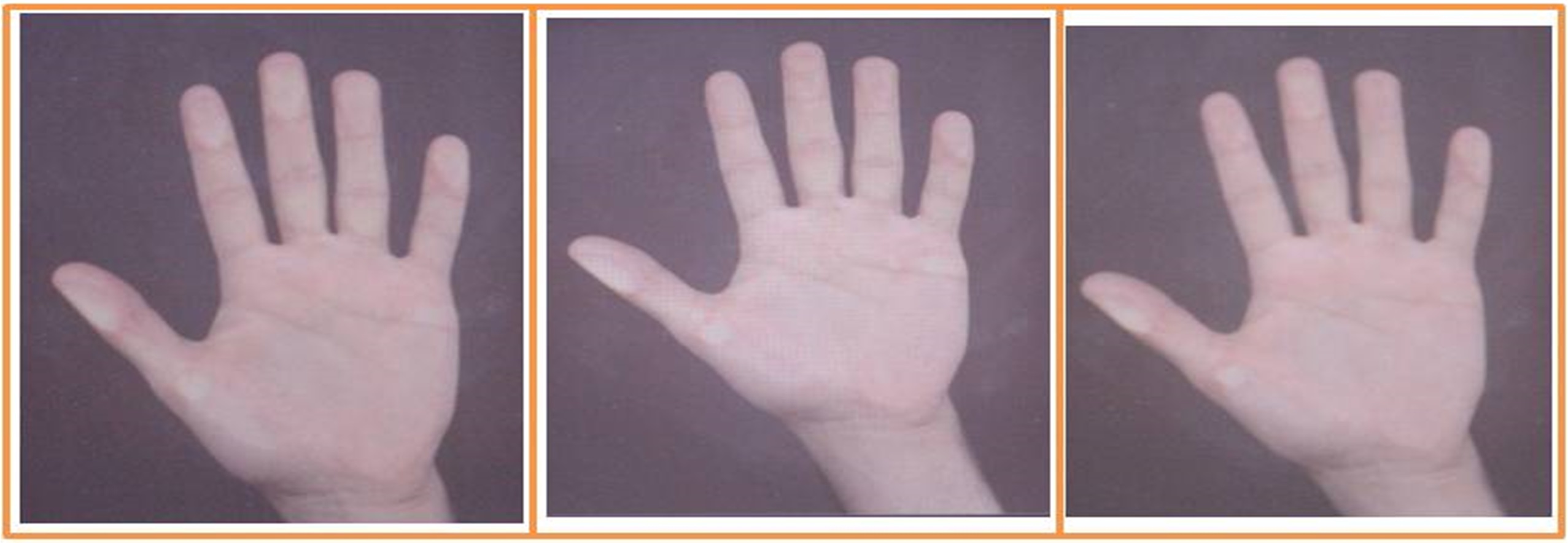}
} 
\hfill  
\subfigure[]{
    \includegraphics[width=0.40\textwidth, height=5.5 cm] {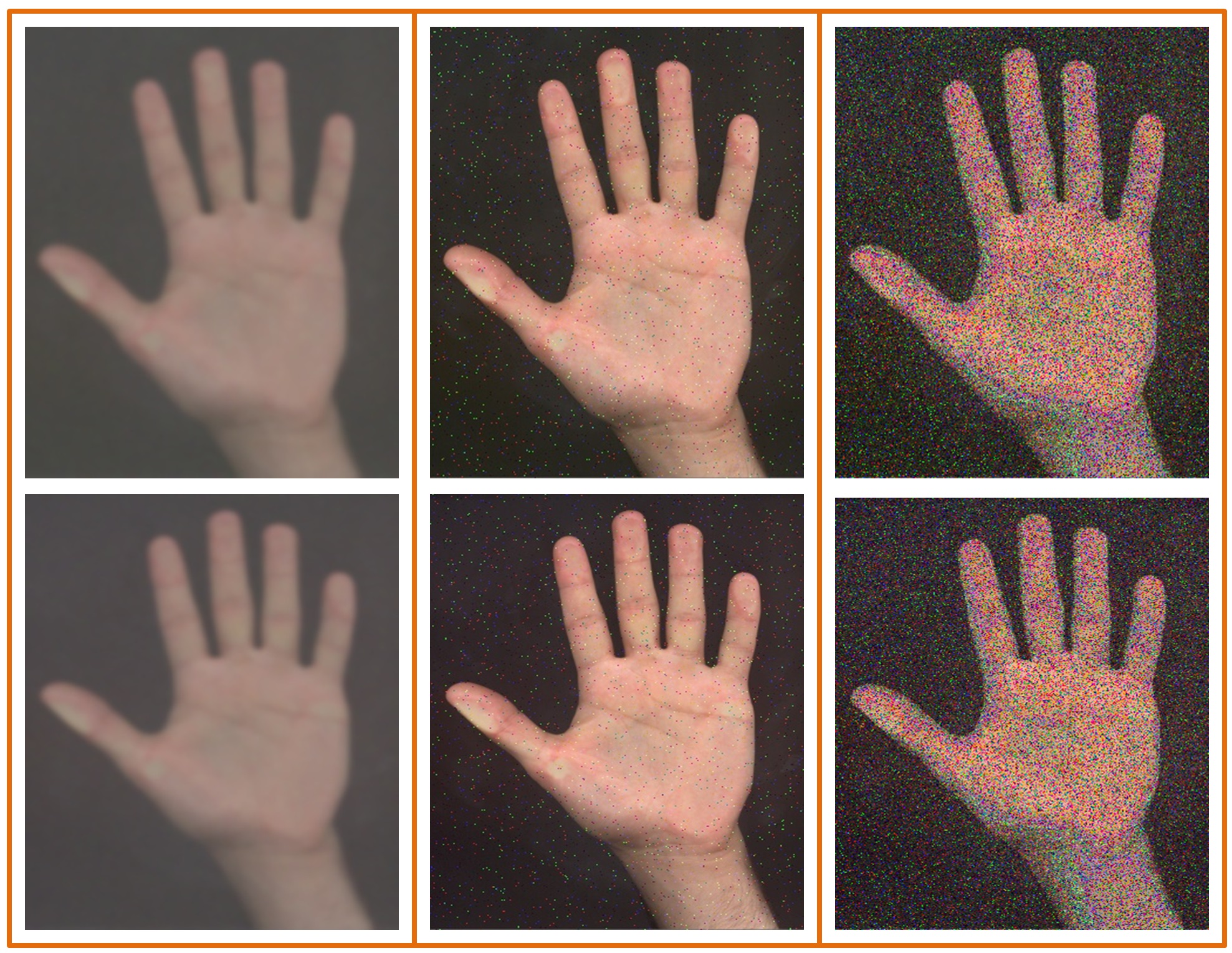}
} 

\hfill  
% figure caption is below the figure
\caption{a) Samples of original hand images of a subject (top-row) which was originally captured by a HP Scanjet 5300 c scanner at 45-dpi. b) Fake hand images with natural degradation acquired from corresponding original hand images (middle-row) using a Canon EOS 700D camera at 72-dpi. c) Fake hand images with artificial noise degradation (bottom-row) by introducing: (left) Gaussian blur, (middle) salt and peeper, and (right) speckle noise. Two fake samples are shown column-wise for each type of added noise. }
\label{fig:2} 
\end{figure}
Biometrics is a secure tool for identity verification in different government, industrial, and forensic applications. The physiological and behavioral biometric modalities have already been established and ameliorated over the decades \textcolor {blue}{\cite{doi2005discrete}}. While biometric authentication renders security against illegitimate identities, it is susceptible to spoofing attacks. In this presentation attack scheme, an imposter replicates a genuine user's  biometric trait(s) to represent a legitimate person. Different spoofing methods (e.g., synthetic, manipulated, or reconstructed trait, etc.) have been attempted on various modalities to undermine the reliability as well as security of the biometric systems \textcolor{blue}{\cite{galbally2013image}}. A biometric system's legitimacy is compromised by different presentation attacks, also known as direct attacks \textcolor{blue}{\cite{chingovska2014biometrics}}. 

As spoofing is a severe challenge to biometrics, necessary countermeasures should be addressed to maintain individualization privacy. The researchers have emphasized various forms of presentation attack detection (PAD) techniques mainly for unimodal and multimodal environments with the face \textcolor{blue}{\cite{chingovska2014biometrics}, \cite{farmanbar2017spoof}}, fingerprint \textcolor{blue}{\cite{nogueira2016fingerprint}, \cite{chugh2018fingerprint}}, palmprint \textcolor{blue}{\cite{bhilare2018study}},  iris \textcolor{blue}{\cite{raghavendra2015robust}}, speech \textcolor{blue}{\cite{wu2017asvspoof}, \cite{korshunov2017impact}}, and other traits. Several hardware and software-specific fundamental techniques are deployed for PAD in the biometric contexts. In the hardware-based method, an expensive sensor is employed for image acquisition from which the constraints about liveness such as the thermal facial map, blood pressure, perspiration, hand-vein temperature, and other salient features are computed \textcolor{blue}{\cite{czajka2013biometric}, \cite{faundez2014new}, \cite{harvey2018characterization}, \cite{travieso2014hand}, \cite{chen2016finger}}. Technological advancements of the sensor devices offer to acquire the thermal \textcolor{blue}{\cite{czajka2013biometric}, \cite{faundez2014new}, \cite{travieso2014hand}, \cite{chen2016finger}, \cite{bartuzi2018thermal}}  %\textcolor{blue}{} [10-13, new ref], 
and hyperspectral \textcolor{blue}{\cite{ferrer2014approach}} images, which direct to devise high-security measures such as liveness detection to discard the fake hand biometric samples. On the contrary, the software-based method discriminates between the real and counterfeit images using a computable characteristic, such as visual quality assessment \textcolor{blue}{\cite{zhang2013edge}, \cite{reenu2013wavelet}, \cite{sun2018spsim}, \cite{wang2004image}, \cite{gao2018image}, \cite{sellahewa2010image}, \cite{guan2017visual}, \cite{martini2012image}}. Software-oriented methods are advantageous than hardware (i.e. sensor-specific) approaches regarding the cost of sensors, and the former can be incorporated easily into a deployed authentication system. In general, the printed photos of a trait (e.g. face \textcolor{blue}{\cite{chingovska2014biometrics}, \cite{farmanbar2017spoof}}), synthetic biometric samples (e.g. silicone fingerprint \textcolor{blue}{\cite{chugh2018fingerprint}}, synthetic speech \textcolor{blue}{\cite{wu2017asvspoof}}, etc.), and electronic screen display (e.g. iris image captured by an iPad \textcolor{blue}{\cite{raghavendra2015robust}}) are established spoofing vulnerabilities. Table 1 summarizes the findings. 

Though hand geometry and other hand-based modalities are  useful for identity verification in industry attendance maintenance, legal and forensic purposes \textcolor{blue}{\cite{bera2014hand}}, it is also exposed to direct attacks \textcolor{blue}{\cite{chen2005fake}}. The spoofing attack detection (SAD) using the thermal hand image has been explored in \textcolor{blue}{\cite{bartuzi2018thermal}}. Individual authentication is not adequate to strengthen the security of a deployed hand biometric system. Also, liveness recognition is a vital task before the actual authentication of a licit user. Hence, there is ample scope to develop a SAD method to deter fraudulent endeavors using hand images before a biometric authentication task, which is the main objective of this paper. 

%#####################################
\begin{table*}  [!htbp] % {0.9\textwidth}
% table caption is above the table
\caption{Common biometric traits for spoofing attacks, and detected with software-based methods}
\label{tab:1}       % Give a unique label
% For LaTeX tables use
  \begin{tabularx}{\linewidth}{
    >{\hsize=1.0\hsize}X% 10% of 4\hsize 
    >{\hsize=0.7\hsize}X% 30% of 4\hsize
    >{\hsize=1.3\hsize}X% 30% of 4\hsize 
       % sum=4.0\hsize for 4 columns
  }

%\hline\noalign{\smallskip}
\toprule
\textbf{Attack method}  &	\textbf{Biometric trait(s)} &	\textbf{Spoof sample creation methodology} \\ 
\noalign{\smallskip}\hline\noalign{\smallskip}
Printed photo 
\textcolor{blue}{\cite{galbally2013image},\cite{chingovska2014biometrics},\cite{farmanbar2017spoof}, \cite{raghavendra2015robust}}  &	Face, iris, fingerprint,
palmprint &	A high-quality printout of the trait is displayed to the sensor. The success of the attack depends on the quality of printed photos.
 \\ \hline
Electronic screen display \textcolor{blue}{\cite{raghavendra2015robust}} &
Iris &	The artifact is produced using a sensor from an enrolled real sample, which is displayed in an electronic display device such as an iPad (4th generation), or a Samsung Galaxy Pad.  \\ \hline
Electronic screen display (\textbf{our work}) &	Hand geometry &	The fake hand dataset with natural degradation is created using a Canon EOS 700D camera from the original sample, which is displayed in a laptop. Another fake hand dataset is created with additional artificial degradation with the Gaussian blur, salt and pepper, and speckle noise.   \\
\noalign{\smallskip}\hline
\end{tabularx}
\end{table*}
\begin{table*}  [!htbp] % {0.9\textwidth}
% table caption is above the table
\caption{Approaches developed for software-based spoofing detection on hand-based modalities}
\label{tab:2}       % Give a unique label
% For LaTeX tables use
  \begin{tabularx}{\linewidth}{
    >{\hsize=0.80\hsize}X% 10% of 4\hsize 
    >{\hsize=0.35\hsize}X% 30% of 4\hsize
    >{\hsize=1.7\hsize}X% 30% of 4\hsize 
       % sum=4.0\hsize for 4 columns
  }
%\hline\noalign{\smallskip}
\toprule
 \textbf{Ref.} &	\textbf{Trait(s)} &	\textbf{Salient feature (under attack)} \\
\noalign{\smallskip}\hline\noalign{\smallskip}
\textcolor{blue}{\cite{bartuzi2018thermal}} & 	Hand geometry &	Thermal features of the hand for presentation attack detection. It employs CNN to determine the real and fake representations of hands imaged in the visible light and thermal spectrum.
 \\ \hline
\textcolor{blue}{\cite{chatterjee2018low}} &	Palmprint &	Fringe projection (Fourier analysis of 3-D texture of palmprint), Biospeckle analysis (sum of pixel-wise absolute value differences among n-images).
 \\ \hline
\textcolor{blue}{\cite{xia2018novel}} &	Fingerprint &	Local Binary Differential Excitation (LBDE) describes amplitude variation among neighbourhood pixels. Local binary gradient orientation (LBGO) describes the orientation information. Weber Local Binary Descriptor (WLBD) is the combination of LBDE and LBGO. 
  \\ \hline
\textcolor{blue}{\cite{nogueira2016fingerprint}}  &	Fingerprint &	Thousands of real and fake fingerprints are tested for liveness using CNN. The local binary pattern (LBP) is employed for feature extraction of a local block.   \\
 \hline
\textcolor{blue}{\cite{farmanbar2017spoof}} & Palmprint and Face  &	Seven-FR-IQMs: mean-square error (MSE), peak signal-to-noise ratio (PSNR), structural similarity index (SSIM), maximum difference (MD), normalized cross-correlation (NCC), normalized absolute error (NAE) and average difference (AD). \\
 \hline
\textcolor{blue}{\cite{qiu2017finger}}  &	Finger-vein &	Total variation regularization, also known as total variation denoising decomposes the original vein images into structure and noise components. Next, LBPs are computed from those two component images. \\
 \hline
\textbf{Ours}  &	Hand geometry &	Ten FR-IQMs: Proposed gradient-based contour profiles similarity metric and existing nine other metrics.  \\

\noalign{\smallskip}
\hline
\end{tabularx}
\end{table*}

Image quality assessment (IQA) is preponderantly applicable in several tasks of image processing, such as image compression, image enhancement, super-resolution, and others \textcolor{blue}{\cite{liu2011image}, \cite{rahul2019fqi}, \cite{8640853}}. IQA is also suitable for software-based spoofing detection using  biometric samples \textcolor{blue}{\cite{galbally2013image}}, \textcolor{blue}{\cite{chingovska2014biometrics}}. Among several standard image quality metrics (IQM), gradient magnitude plays a vital role in describing the visual quality of generic images \textcolor{blue}{\cite{liu2011image}, \cite{bondzulic2018gradient}}. It is based on intensity variations in the edge-maps. This metric has been addressed in various state-of-the-art approaches to discriminate between the original and distorted images \textcolor{blue}{\cite {6678238}}.

In this work, we have considered a threshold parameter to offer flexibility in gradient computation. The main advantage of this modification in gradient computation with thresholding is to consider the extent of intensity difference between neighborhood pixels. This paper presents the modified gradient similarity i.e., threshold-based gradient magnitude similarity (GMS) for quality assessment in the context of anti-spoofing on hand biometrics. %
The proposed framework follows an electronic screen display based anti-spoofing technique to hinder the susceptibility of attack using a fake hand, which is presented to a deployed verification system. This advanced system is pictorially ideated in Fig.\ref{fig:main}. 

For experimentation, a fake hand dataset with natural degradation is created using a Canon EOS 700D camera from 255 subjects from the Bogazici University (BU) hand dataset \textcolor{blue}{\cite{yoruk2006shape}, \cite{dutagaci2008comparative}}. Three original and corresponding fake samples of an individual are shown in Fig.\ref{fig:2}(a-b). Moreover, we have also created a fake dataset comprising the same population with artificial degradation by introducing three types of noises, namely, Gaussian blur, salt and pepper, and speckle noise. Samples of degraded hand images with these noises are shown in Fig.\ref{fig:2}.c. Ten full reference quality measures are computed for experiments using machine learning and deep learning techniques. 

The contributions of this paper are summarized as follows:
\begin{itemize}
    \item A new visual quality metric based on the gradient magnitude similarity is proposed. Total ten quality metrics are computed to distinguish an original hand image from a fake sample.  
    \item Two types of fake hand datasets are created with natural and artificial degradation for spoofing detection using the original images from the Bogazici University hand dataset.
    \item The proposed method is evaluated using both conventional and deep learning methods for spoofing detection with various experimental scenarios.
    \item The experimental results signify that presentation attack detection using quality assessment can be effective for a hand biometric authentication system. This proposed method can reduce the susceptibility of a hand biometric system regarding spoofing detection.
\end{itemize}

The rest of this paper is organized as follows. Section 2 describes a study on the related works on PAD. Section 3 provides a brief on the IQM. Section 4 presents the experimental results, and Section 5 draws the conclusion. 

%############################################
\section{Related Study}
\label{study}
A general convention is believed that fake samples differ from real images as different sensors are employed in either case of image acquisition. The quality of an image sample should differ from another sample that is collected by a different sensor. The quality difference may include intensity, luminance, color, blurriness, and other variations. The hypothesis of ‘quality-difference’ between the real and spoofed samples is justified in \textcolor{blue}{\cite{galbally2013image}}. Their method is evaluated using generic full-reference (FR) and no-reference (NR) quality estimation using features computed from printed-photos of the face, fingerprint, and iris samples. In \textcolor{blue}{\cite{banitalebi2019image}}, an FR wavelet-based IQA is tested. A reduced reference (RR) based quality evaluation method, namely feature quality index, is presented in \textcolor{blue}{\cite{rahul2019fqi}}. A blind and training-free image blur estimation metric based on the spatial data is proposed in \textcolor{blue}{\cite {bong2014blind}}. The algorithm is developed using the double Gaussian convolution. This method is highly correlated to human perception of blurriness. The quality difference between the real and counterfeit images plays a key role for PAD. It has been studied that mainly the face \textcolor{blue}{\cite{jia2020face}}, \textcolor{blue}{\cite{pinto2020leveraging}}, \textcolor{blue}{\cite{fourati2020anti}}, and finger-print \textcolor{blue}{\cite{tolosana2019biometric}} have received much research attention for PAD. In \textcolor{blue}{\cite{jia2020face}}, PAD based on face images acquired using mobile devices is proposed. Also, IQA based on the face makeup is proposed in \textcolor{blue}{\cite{rathgeb2020vulnerability}}.

A spoofing detection method using the dorsal hand vein images is presented in \textcolor{blue}{\cite {patil2016assessing}}. It is anticipated that multimodal systems are intrinsically robust against spoofed artifacts. The anti-spoofing method is also validated for multibiometrics \textcolor{blue}{\cite{chingovska2014biometrics}}, \textcolor{blue}{\cite{biggio2016statistical}}, and fusion strategies \textcolor{blue}{\cite{korshunov2017impact}}. Sajjad et al. \textcolor{blue}{\cite {sajjad2019cnn}} have proposed a multimodal biometric system using the conventional hand-crafted features of the fingerprint, palmprint and face modalities. It is also useful for spoofing detection using deep convolutional neural networks (CNN) based high-level features. The GoogLeNet is used as a backbone to compute deep features for spoofing detection. Another mode of spoofing attack using the electronic screen display is illustrated  in \textcolor{blue}{\cite{raghavendra2015robust}}. In their proposal, a real sample is displayed as a printed copy to a Smartphone or other electronic devices. A fake sample is generated by acquiring an image of a real sample with a camera or Smartphone. This framework has been tested over the various iris datasets, rendering satisfactory performance.  In a similar direction, we have applied a light-weight MobileNetV2 \textcolor{blue}{\cite{sandler2018mobilenetv2}} deep model for our baseline evaluation for hand biometric spoofing detection using our fake datasets.

Hand geometry is a well-known biometric mode for human verification \textcolor{blue}{\cite{bera2019finger}}. Klonowski et al. \textcolor{blue}{\cite {klonowski2018user}} have proposed an algorithm for hand geometry and crookedness identification. To locate the region of interest, the convex hull is employed. The feature set includes the geometric measurements and crookedness of the fingers. Barra et al. \textcolor{blue}{\cite {barra2019hand}} have proposed a hand biometric system in the Android mobile environment. Hand segmentation process is based on the detection of the convexities and concavities of the fingers. Four types of features are computed from the hand-shape related to the length, area, angle, and ratio. A decidability metric is defined for feature selection to improve the performance. Another hand segmentation approach from cluttered background is described in \textcolor{blue}{\cite {bapat2017segmentation}}. In \textcolor{blue}{\cite {Bera8032481}}, a forward-backward feature selection method is applied to improve finger biometric authentication accuracy. In \textcolor{blue}{\cite{jaswal2019multimodal}}, a  feature-level fusion method using hand shape, geometry, and palm print features is performed. Decision-level fusion using the hand geometric feature is presented in \textcolor{blue}{\cite{bera2015fusion}}. On the other research line, though, hand-based modalities such as fingerprint, palmprint, and finger-vein have been tested for PAD. However, to the best of our knowledge, only a few comprehensive works exist using hand geometry in the direction of spoofing attack detection \textcolor{blue}{\cite{chen2005fake}, \cite {bartuzi2018thermal}}. An exemplary study on anti-spoofing, especially on hand-based modalities, is presented in Table \ref{tab:2}.

An easy spoofing method is ideated with the hand modality in \textcolor{blue}{\cite{chen2005fake}}. The fake hands are created using the plaster and making a hand silhouette with a paper card. The fake hand (made by plaster or paper card) is presented to the hand geometry verification system for illegitimate access. This presentation attack has been tested with the fake hands of only five different persons. However, no suitable experimental description has been illustrated.  Bartuzi and Trokielewicz \textcolor{blue}{\cite {bartuzi2018thermal}} have proposed a method for PAD using thermal hand images. To create the fake hand dataset, the printed copy of the original hand images is used. Two CNN backbones, namely the AlexNet and VGG-19 models are used for deep feature extraction. Inspired by these approaches, we have developed an anti-spoofing technique using the visual qualities of images in the context of hand biometrics. We have tested our fake sample datasets with conventional and deep learning classification schemes. 

\section{Image Quality Metrics (IQM)}
\label{IQM}
The visual quality of an image entails several fundamental characteristics of the image itself rather than its context. The generic attributes of an image are very significant for quality assessment \textcolor{blue}{\cite{liu2011image}}, with applications in computer vision and image processing. Image quality measurement can be classified as subjective or objective \textcolor{blue}{\cite{bong2014blind}}. In subjective evaluation, a human is involved in measuring the visual quality (i.e. blurriness assessment), maintaining the subjective evaluation standards such as ITU-R BT.500. On the contrary, objective assessment methods are generally based on the transformed (e.g., wavelet domain) and spatial (e.g., structural similarity) domains. Many of the objective estimations are mainly full-reference (FR), reduced-reference (RR), and no-reference (NR) methods. Different types of features are established for quality assessment of generic and unique images, such as biometric samples. 

\subsection{\textbf{Related Image Quality Metrics}}
The available IQA approaches are the full-reference (FR), reduced-reference (RR), and no-reference (NR) method \textcolor{blue}{\cite{wang2004image}, \cite{gao2018image}, \cite{rahul2019fqi}}. In the FR method, a clear image is considered as a reference with respect to which distorted image's quality is assessed. Incomplete information about a reference image is available in the RR method. However, in the NR method, the quality is determined in the absence of a reference image. As a result, the NR method is more challenging than the FR method regarding the quality calculation. Here, an FR method containing ten quality metrics is described based on the sensitivity measures such as the noise, structural similarities, gradient magnitude, wavelets, and others. The quality is measured according to the method, as described in \textcolor{blue}{\cite{galbally2013image}}. Also, five IQMs (indexed with (a)-(e)) are discussed next, and those metrics are also assessed in their work.

\begin{figure*}
\centering
% Use the relevant command to insert your figure file.
% For example, with the graphicx package use
\subfigure []{
  \includegraphics[width=0.9\textwidth, height=4.0cm] {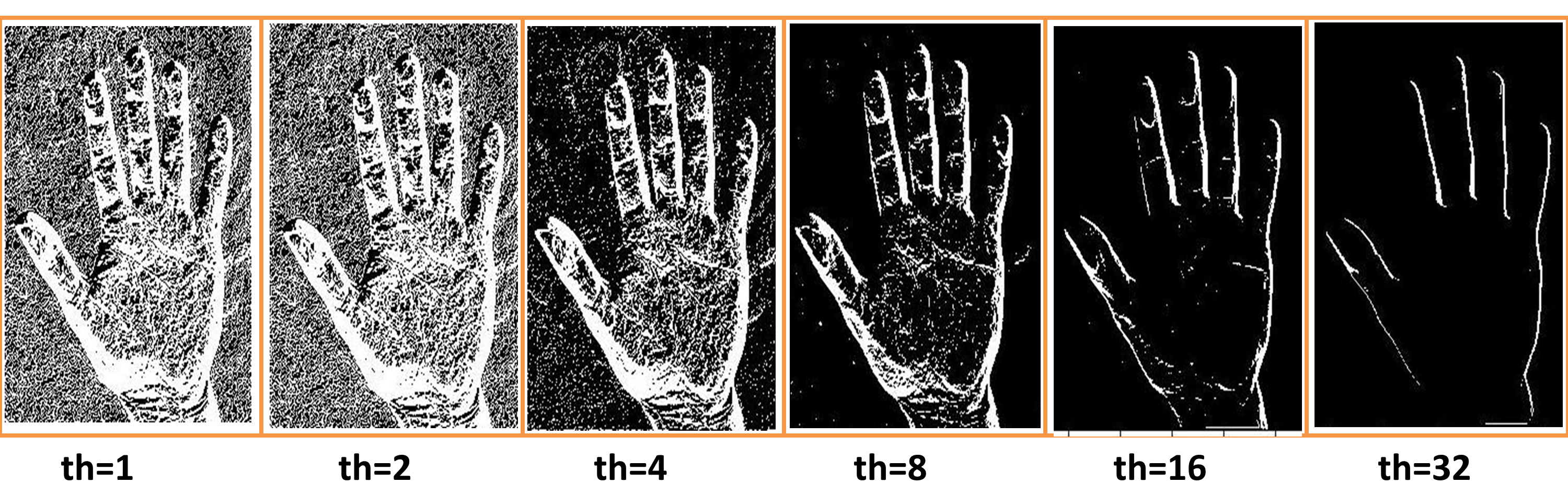}
} 

\subfigure[]{
    \includegraphics[width=0.9\textwidth, height=4.0cm] {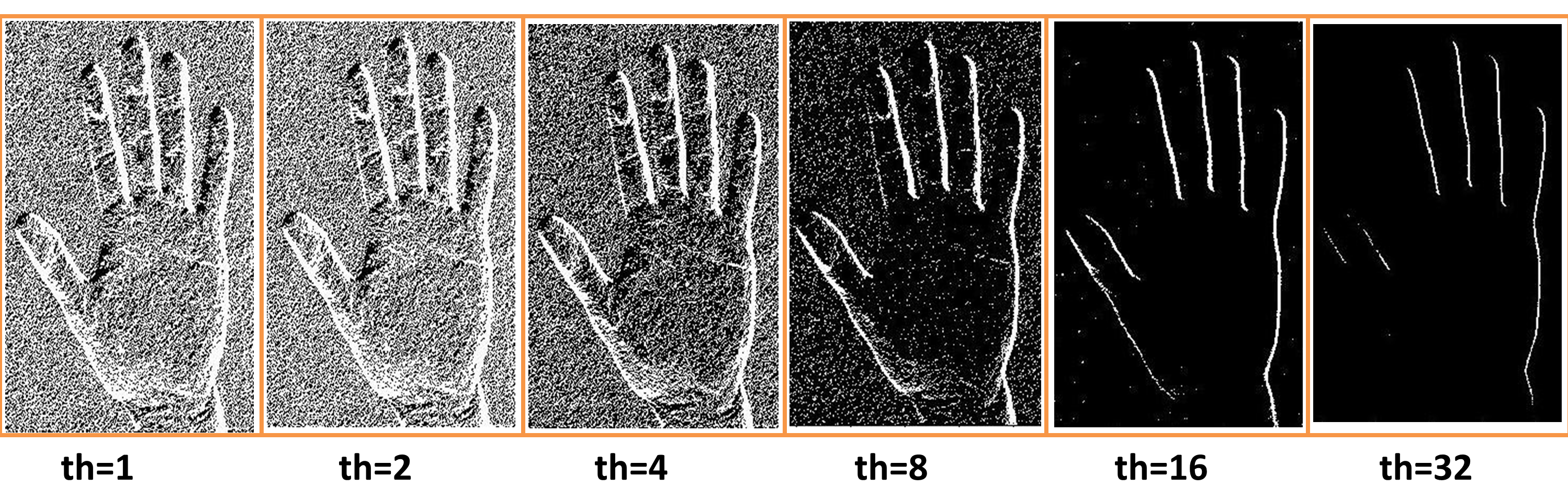}
} 
\hfill    
% figure caption is below the figure
\caption{ Binary images of the first images (column 1 in Fig.\ref{fig:2}.a-b) rendered with the variations of thresholds, for p=1,2,3,4, and 5. (a) Variations in the real hand images obtained from eq.13, (top row). b) Deviations of the respective spoofed hand images (bottom row). }
\label{fig:real_th}       % Give a unique label
\end{figure*}
Firstly, an input color image is converted into a grayscale image I with $N$ rows and $M$ columns of pixels, and it is denoted as $I_{N\times M}$. A pixel is indexed by $(i, j)\in \mathbb{R}^2$ with the row-index $i=1, 2, …, N$ and column-index $j=1, 2,…,M$. Here, $I_{N\times M}$ is smoothed with a $3\times 3$  low-pass Gaussian filter with the correlation kernel (standard deviation $\sigma =0.5$), which produces a distorted image $\bar{I}_{N\times M}$. The pixels which lie outside the bound are replicated with the values of boundary pixels. The quality difference ($\Delta$) between the original and distorted images is measured with respect to a metric one at a time, defined below: 
\begin{equation}
\Delta(I, \bar{I}) = IQM(I) - IQM(\bar{I})	
\end{equation}

\noindent \textbf{Mean Square Error (MSE):} MSE is a simple and mostly used metric for IQA due to its mathematical simplicity. It is based on pixel-wise intensity differences:
\begin{equation}
MSE(I, \bar{I})= \mathop{\sum^{N}\sum^{M}}_{i=1\ j=1}\big( I_{i,j}-\bar{I}_{i,j} \big)^2
\end{equation}

\noindent \textbf{Peak Signal to Noise Ratio (PSNR):} PSNR is derived from the squared differences between pixel-wise intensities, and its underlying basis is MSE. It is attractive due to its simple mathematical easiness like MSE. Both the MSE and PSNR do not consider the quality of visual information:
\begin{equation}
PSNR(I, \bar{I})= 10\log_{10} \Big(max(I^2)\big/ MSE(I, \bar{I})\Big) 
\end{equation}

\noindent \textbf{Structural Content (SC):} SC is another pixel-based metric that computes the ratio between the original and distorted images:
\begin{equation}
SC(I, \bar{I})= \mathop{\sum^{N}\sum^{M}}_{i=1\   j=1} I_{i,j}^2\Big/ \mathop{\sum^{N}\sum^{M}}_{i=1\   j=1} \bar{I}_{i,j}^2
\end{equation}

Edges and corners represent significant information about an image. These are apposite for quality assessment too. The edge and corner differences are computed between I and $\bar{I}$.\\
\noindent \textbf{Edge Difference (ED):} The Sobel operator is applied to determine binary edge maps from which a total edge difference (absolute value) is calculated. Here, the edges are denoted with E and Ē which are computed from I and $\bar{I}$, respectively:
\begin{equation}
ED(I, \bar{I})= \frac{1}{NM} \mathop{\sum^{N}\sum^{M}}_{i=1\ j=1} \big| E_{i,j} - \bar{E}_{i,j}  \big|
\end{equation}

\noindent \textbf{Corner Difference (CD):} The Harris corner detector is followed to compute the number of corners $N_{CR}$ and $\bar{N}_{CR}$ from binary images of real and distorted images, respectively. It is simply delineated as a ratio related to the number of corners detected in I and $\bar{I}$:   
\begin{equation}
CD(I, \bar{I})= \frac{\big| N_{CR} - \bar{N}_{CR} \big| }{ max \Big( N_{CR}, \bar{N}_{CR} \Big) } 
\end{equation}

\noindent \textbf{Entropy Difference (EyD):} Entropy, $H$, is a statistical measure to define a grayscale image's randomness. It is defined as $H=-\sum p log_2(p)$ , where p represents the histogram counts regarding  the image's grayscale intensities. This metric computes the entropy difference between I and $\bar{I}$:
\begin{equation}
EyD(I, \bar{I})= \frac{\big| H(I) - H(\bar{I}) \big| }{ max \Big( H(I), H(\bar{I}) \Big) } 
\end{equation}

\noindent \textbf{Structural Similarity Index Measure (SSIM):} The similarity index measure relies on the luminance (\textit{l}), contrast (\textit{c}), and structural (\textit{s}) perceptual qualities of the human visual system \textcolor{blue}{\cite{wang2004image}}. The mean SSIM is calculated for estimating the overall perceived image quality.  The most straightforward formulation is given as:
\begin{equation}
%\begin{aligned}
SSIM(I, \bar{I})= \big( \textit{l}(I, {\bar{I}})^\alpha \big).\big( \textit{c}(I, {\bar{I}})^{\beta} \big).\big( \textit{s}(I, {\bar{I}})^\gamma  \big)
%\end{aligned}
\end{equation}

where $ \textit{l}(I, \bar{I})= \frac{2 \mu_I\mu_{\bar{I}} + \epsilon  }{\mu_I^2 + \mu_{\bar{I}}^2 + \epsilon } $, 
$ \textit{c}(I, \bar{I})= \frac{2 \sigma_I\sigma_{\bar{I}} + \epsilon  }{\sigma_I^2 + \sigma_{\bar{I}}^2 + \epsilon } $, 
$ \textit{s}(I, \bar{I})= \frac{\sigma_{I{\bar{I}}} + \epsilon  }{\sigma_I\sigma_{\bar{I}} + \epsilon }$
and $\alpha$, $\beta$, and $\gamma$ are positive constants for emphasizing each component relatively. The mean of $I$ $({\mu}_I)$, the mean of $\bar{I}$  ($\mu_{\bar{I}}$), the variance of $I$ ($\sigma_I$), the variance of $\bar{I}$ ($\sigma_{\bar{I}} $), and the co-variance of $I$ and ${\bar{I}}$ ($\sigma_{I\bar{I}}$) are vital parameters to determine the quality of real ($I$) and distorted (${\bar{I}}$) images, which are considered as a set of the pixel block. A small positive constant ${\epsilon}$ is considered to make the denominator non-zero. The SSIM estimates the quality of each pixel with reasonable accuracy. Its mathematical foundation is computationally efficient and appealing for quality analysis of generic images. It is a popular metric, and  several variations have been developed for IQA, such as super-pixel structural similarity (SPSIM) \textcolor{blue}{\cite{sun2018spsim}}, \textcolor{blue}{\cite{8640853}}.

\noindent \textbf{Edge-Strength Similarity-Based Image Quality Metric (ESSIM):} ESSIM represents the semantic information as the edge-strength of an object regarding each pixel \textcolor{blue}{\cite{zhang2013edge}}. It computes whether a pixel belongs to the edge of a semantic object. The ESSIM is measured regarding horizontal, vertical, and diagonal directions. The directional derivatives are used to determine the directional edge strength using a suitable Scharr kernel. The ESSIM estimates the visual fidelity between a reference and distorted image regarding the edge-strength map:
\begin{equation}
ESSIM(I, \bar{I})= \frac{1}{NM}\mathop{\sum^{N}\sum^{M}}_{i=1\ j=1} \frac{2e(I_{i,j})e(\bar{I}_{i,j})+ \epsilon}{(e(I_{i,j})^2 + (e(\bar{I}_{i,j})^2 + \epsilon}
\end{equation}
where e() computes the edge-strength of corresponding pixels indexed with (i, j), and $\epsilon$ is a small positive constant as defined earlier. \\   

\noindent \textbf{WAvelet based SHarp features (WASH):} In WASH, the perceived quality is based on sharpness, and zero-crossings in the wavelet domain \textcolor{blue}{\cite{reenu2013wavelet}}. The sharpness of an image is estimated from the energy in the wavelet sub-bands. The wavelet-based zero-crossings follows the Laplacian method for edge detection using the second-order derivative of an image. Note that it follows successive operations to compute WASH (W). The similarity ($\upsilon$) metric derived using the sharpness ($\lambda$) of a reference, and  the distorted image is defined as:

\begin{equation}
%\begin{align}
\upsilon(I, \bar{I})= \frac{2I_{\lambda}\bar{I}_{\lambda}+ \epsilon}{I_{\lambda}^2 + \bar{I}_{\lambda}^2 + \epsilon}
%\end{align}
\end{equation}
where $\epsilon$ is defined above. The zero-crossing is defined using the edge structural similarity ($E_S$) of the reference and distorted images in three wavelet sub-bands (namely, the $W_{LH}$, $W_{HL}$, and $W_{HH}$): 
\begin{equation}
%\begin{aligne}
Z(I, \bar{I})= \Pi_{X}E_{S}(I, \bar{I})
%\end{equation}
\quad \textrm{where} 
%\begin{equation}
{\Pi}_{X}E_{S}=\frac{\Sigma \left( NE_I \cap NE_{\bar{I}} \right)}{\sqrt{\Sigma NE_{I}}\sqrt{\Sigma NE_{\bar{I}}}}
%\end{aligne}
\end{equation}
where $X$ denotes the sub-bands of single-level wavelet decomposition of the image. Finally, the WASH is measured as $W(I, \bar{I})= \upsilon(I, \bar{I})^\gamma + Z(I, \bar{I})^{(1-\gamma)}$  with $\gamma=0.8$ for a higher weightage to sharpness.

\subsection{\textbf{Proposed Gradient Magnitude Similarity (GMS)}}
Gradient conveys important visual information for IQA 
\textcolor{blue}{\cite{liu2011image}, \cite{bondzulic2018gradient}}. It reflects structural and contrasts differences of an image. Generally, a small edge discontinuity due to illumination imperfection or manipulated artifact can be accounted for gradient operators, which could not be reckoned only from a grayscale image. 
The general idea to compute the gradient is to apply convolution between a given image with a linear kernel operator. Commonly used filters include the Prewitt, Sobel, Scharr filters. In \textcolor{blue}{\cite{6678238}}, the Prewitt filters along the horizontal ($x$) and vertical ($y$) directions are used for gradient computation. For example, Prewitt's $3 \times 3$ operators for gradient computation are given below.

\vspace{0.3 cm}
\begin{minipage}{0.2\textwidth}
\begin{tabular}{|c|c|c|}
\hline
 1 & 0 & -1 \\
\hline
 1 & 0 & -1 \\
\hline 
 1 & 0 & -1 \\
\hline
\end{tabular}
\end{minipage}
\begin{minipage}{0.2\textwidth}
\begin{tabular}{|c|c|c|}
\hline
 1 & 1 & 1 \\
\hline
 0 & 0 & 0  \\
\hline 
 -1 & -1 & -1 \\
 \hline
\end{tabular}
\end{minipage}

\vspace{0.5 cm}
\noindent The convolution ($\otimes$) is applied to the input image I(i,j) with these operators:

\begin{equation}
G_x(i,j)= I(i,j) \otimes h_x 
\quad \textrm{and} \quad
G_y(i,j)= I(i,j) \otimes h_y 
\end{equation}

We have proposed a simple method to compute the gradient values ( $G_x$ and $G_y$) by considering the local intensity difference between two adjacent pixels at the location I(i, j) and I(i, j+1) along the X-axis. The 1-D gradient operator $h_x=[1, 0, -1]$ is simplified as $h_x=[1, -1]$. Most importantly, a control parameter \textit{th} is introduced to eliminate trivial intensity profile variations between consecutive pixels in the gradient computation. The modified 1-D masks are given as $G_x=\frac{1}{th}[ 1, -1 ]$, and $G_y=\frac{1}{th}[ 1, -1 ]^T $. $G_x$ and $G_y$ are akin to traditional gradient magnitude  $(G_M\approx G_x+G_y)$ when \textit{th}=1. In our proposed method, the gradient maps are computed in a pixel-wise manner, defined as follows:
\begin{equation}
G_x(i,j)= \frac{1}{th} \big| I(i,j) - I(i,j+1)  \big|
\end{equation}

\begin{equation}
G_y(i,j)= \frac{1}{th} \big| I(i,j) - I(i+1,j)  \big|
\end{equation}

\begin{equation}
  \Delta G_x= \mathop{\sum^{N}\sum^{M}}_{i=1\ j=1}\big( G_x(i,j)-\bar{G}_x(i,j) \big)^2
\end{equation}

\begin{equation}
\Delta G_y= \mathop{\sum^{N}\sum^{M}}_{i=1\ j=1}\big( G_y(i,j)-\bar{G}_y(i,j) \big)^2
\end{equation}

Similarly, $\bar{G}_x$ and $\bar{G}_y$ are defined for $\bar{I}$ in the same manner. A higher value of \textit{th} is considered for a significant variation from high- to low-intensity between neighborhood pixels. It is defined as $\textit{th}=2^p$ where p=0, 1, …, 5. The significance of changing \textit{th} to determine the edge-strength is illustrated in Fig.\ref{fig:real_th}.  
The variations of intensities are presented in the binary images for both types of images. This method is efficient for  geometric feature extraction based on contour profiles of the real hand images. It is noted that for a higher \textit{th}, the intensity profile variations in gradient magnitudes are remarkable \textcolor{blue}{\cite{bera2017finger}, \cite{Bera8032481}}. This important modification in the gradient map computation facilitates the formulation of the proposed gradient similarity metric, and the influence of \textit{th} in $G_M$ is described in the experiments. In addition, normalizing the overall quality with the average factor $1/NM$ can be considered another modification in the proposed mean GMS map (eq.17-18). The average value considers the equal importance of each pixel to compute the image quality. Therefore, it computes the average gradient magnitude similarity value to estimate the image quality as:
\begin{equation}
GMS(I, \bar{I})= \frac{1}{NM} \mathop{\sum^{N}\sum^{M}}_{i=1\ j=1} \sqrt{\Delta G_x + \Delta G_y}
\end{equation}
Alternatively, GMS can be defined as:
\begin{equation}
GMS(I, \bar{I})= \frac{2 \Delta G_x \Delta G_y +\epsilon }{NM (\Delta {G_x}^2 + \Delta {G_y}^2 +\epsilon)} 
\end{equation}
where a positive constant $\epsilon=10^{-5}$. 

As a summary, the novelty in formulating the mean GMS metric is thresholding on the gradient computation method using a control parameter \textit{th}. All these aforesaid ten IQMs are used for quality assessment between the original and fake hand images, which are described in the next section.

\section{Experimental Descriptions}
The experiments are conducted for classifying the real and spoofed hand images of 255 persons of the BU dataset. Three original left-hand images per person (i.e., a total of 765 images) are considered for fake datasets creation. The fake hand datasets are created with two variations: (i) Fake hand dataset with natural degradation and (ii) Fake hand dataset with artificial degradation. A comprehensive description of various experiments with these fake image sets is summarized. 
\subsection{Fake Hand Dataset Creation }
The original left-hand images (45 dpi) of 255 subjects of the BU database are chosen as the real hand from which the fake samples are captured. The characteristics of the BU database are described in \textcolor{blue}{\cite{yoruk2006shape}, \cite{dutagaci2008comparative}}. 
\subsubsection{Fake hand dataset with natural degradation}
The original dataset contains three color hand images per person, which are displayed on a laptop one at a time. An image is then acquired as the fake sample (72 dpi) for each real hand using a Canon EOS 700D camera. Fig.\ref{fig:2}.a shows three real hands of a subject, and its corresponding fake samples are shown in Fig.\ref{fig:2}.b. The idea of creating such a fake sample is that the original image of the BU dataset is displayed on a laptop screen, and then an image is captured using the camera from the laptop screen. There is a fixed distance between the camera lens and  the laptop screen. We have maintained the same distance for each capture. The lighting conditions, camera angle, and other imaging factors remain consistent for creating a sample fake dataset. These fake images are distorted naturally.  Due to surrounding environmental conditions during image acquisition, natural artifacts are introduced in the fake samples. Therefore, this spoofing image dataset collection is denoted as fake hand dataset with natural degradation  and it consists of a total of 765 fake samples. 

\subsubsection{Fake hand dataset with artificial degradation}
Our next fake dataset is created by introducing various artificial noises to the same original hand images and denoted as fake hand dataset with artificial degradation. We have added three common noises, namely, Gaussian blur, salt and pepper noise, and speckle noise, to the real images. The samples  are shown in Fig\ref{fig:2}.c. 
We have created one fake sample from each real sample for each category of artificial noise. Thus, this fake hand dataset comprises  \textit{the number of original images per hand}  $\times $ \textit{total number of subjects} $\times$ \textit{ the number of artificial noises} = 3 $\times$ 255 $\times$ 3 = 2295 spoofed images altogether. 

In summary, we have experimented on 765 original images and a total of 765+2295=3060 fake hand images from both fake datasets. All the images are converted into grayscale and resized to $ 400 \times 300 $ pixels, before computing the defined IQMs. 

\begin{table*}
\centering
  \caption{HTERs (\%) estimation using the defined IQMs. The SSIM performs the best results. The ESSIM and proposed GMS have achieved comparable results to SSIM. The best results among individual IQM, and the proposed GMS using different  classifiers are marked in \textbf{bold} font.  }
  \label{tab:commands}
  \begin{tabular}{l| c c c | c c c| c c c | c c c}
    \toprule
    %\multirow{2}{*}{Dataset} &
      \multicolumn{1}{c} { } &
      \multicolumn{3}{c}{\textbf{k-NN}}&
      \multicolumn{3}{c}{\textbf{RF}} &
      \multicolumn{3}{c}{\textbf{SVM (linear)}}&
      \multicolumn{3}{c}{\textbf{SVM (RBF)}}\\
      IQM    & {FFR}	& {FGR} &	{HTER} & {FFR}	& {FGR} &	{HTER} & {FFR}	& {FGR} &	{HTER} & {FFR}	& {FGR} &	{HTER}  \\
      \midrule
        (a) MSE &	35.33 &	35.33&	35.33&	20&	44 &	32 &	23.3 &	33.33 &	28.33&	23.3&	33.33&	28.33\\
       (b) PSNR &	35.33 &	35.33 &	35.33 &	30 &	34.67 &	32.33 &	24.67 &	30 &	27.33 &	24 &	31.33 &	27.67  \\ 
     (c) SC	& 38 &	22 &	30 &	32 &	24 &	28&	23.33 &	42.67 &	33 &	44.67 &	10.67&	27.67 \\
      (d) ED&	37.33&	38&	37.67&	 34.67&	33.33&	34&	56.67&	20&	38.33&	27.33&	34&	30.67\\
    (e) CD	& 46&	58.67&	52.33&	 61.33&	32.67&	47&	68.67&	21.33&	45&	63.33&	24&	43.67\\
    (f) EyD &	40&	41.33&	40.67&	38&	38&	38&	10&	71.33&	40.67&	24.67&	46&	35.33\\
     (g) \textbf {SSIM} &	\textbf {0}&	\textbf {2}&\textbf {	1}	&\textbf {0}&	\textbf {1.33}& \textbf {	0.67}&	\textbf {1.33}&	\textbf {0.67}&	\textbf {1}&	\textbf {0.67}&	\textbf {0.67 }&	\textbf {0.67}    \\
     (h) ESSIM &	2.67&	2&	2.33&	0.67&	1.33&	1 &	1.33&	0.67&	1&	1.33&	0.67&	1 \\
    (i) WASH &	6&	1.33&	3.67&	5.33&	1.33&	3.33&	3.33&	10&	6.67&	6&	2.67&	4.33\\ \hline
    (j) \textbf{Proposed GMS(th=8)} &	\textbf {0} &	\textbf {2.67}&	\textbf {1.33}&	\textbf {0} &	\textbf {2.67}&	\textbf {1.33} &	1.33&	2&	1.67&	1.33&	2&	1.67     \\
    
    \bottomrule
  \end{tabular}
  \label{table:HTER_IQM}
\end{table*}

\begin{table*}
\centering
  \caption{Experiments to determine threshold using two definitions of Gradient Magnitude Similarity (GMS). Top row-set:   GMS  eq. (17), bottom row-set:   GMS  eq. (18). The best results achieved using threshold variation (\textit{th}=8) in both equations are marked with \textbf{bold} font.}
  \label{tab:commands}
  \begin{tabular}{l| c c c | c c c| c c c | c c c}
    \toprule
    %\multirow{2}{*}{Dataset} &
      \multicolumn{1}{c} { } &
      \multicolumn{3}{c}{\textbf{k-NN}}&
      \multicolumn{3}{c}{\textbf{RF}} &
      \multicolumn{3}{c}{\textbf{SVM (linear)}}&
      \multicolumn{3}{c}{\textbf{SVM (RBF)}}\\
      \textit{th}    & {FFR}	& {FGR} &	{HTER} & {FFR}	& {FGR} &	{HTER} & {FFR}	& {FGR} &	{HTER} & {FFR}	& {FGR} &	{HTER}  \\
      \midrule
        1 &	18&	20.67&	19.33&	16.67&	18&	17.33&	4.67&	28&	16.33&	14.67&	12.67&	13.67\\
        2&	10&	15.33&	12.67&	12&	10&	11&	0.67&	28&	14.33&	10.67&	8.67&	9.67\\
        4&	1.33&	4.67&	3&	1.33&	2.67&	2&	0.67&	12.67&	6.67&	4&	2.67&	3.33\\
       \textbf { 8} &\textbf {	0}&	\textbf {2.67}&	\textbf {1.33}&	\textbf {0.67}&	\textbf {2}&	\textbf {1.33}&	\textbf {1.33}&	\textbf {2}&	\textbf {1.67}&	\textbf {1.33}&	\textbf {2}&	\textbf {1.67}\\
        16&	23.33&	26.67&	25&	22&	20.66&	21.33&	12&	26&	19&	8&	27.33&	17.67\\
        32&	39.33&	48&	43.67&	36&	47.33&	41.67&	38.67&	58.67&	48.67&	32&	40.67&	36.33  \\
      \midrule
      1&	39.33&	43.3&	41.33&	35.33&	40&	37.67&	27.33&	32.67&	30&	36&	21.33&	28.67 \\
   2&	37.33&	36.67&	37&	34.67&	33.33&	34&	35.33&	24.67&	30&	34&	27.33&	30.67 \\
4&	31.33&	44.67&	38&	26.67&	43.33&	35&	32&	23.33&	27.67&	18.67&	39.33&	29 \\
8&	33.33&	45.33&	39.33&	22.67&	46&	34.33&	33.33&	26&	29.67&	35.33&	24.67&	30 \\
16&	42.67&	46&	44.33&	37.33&	45.33&	41.33&	32&	54.67&	43.33&	22&	52.67&	37.33 \\
32&	47.33&	51.33&	49.33&	36.67&	54&	45.33&	42&	48&	45&	46.67&	42.67&	44.67 \\
    \bottomrule
  \end{tabular}
  \label{table:GMS}
\end{table*}

\subsection{Experimental Analysis}
Our proposed work is experimented with the hand-crafted features using conventional machine learning techniques and the deep features using CNN. For the latter experiments, MobileNetV2 \textcolor{blue}{\cite{sandler2018mobilenetv2}} is used as a base network. Thus, our experimental setup is two-fold.

\subsubsection{Experimental Results}

At first, we have experimented with naturally degraded fake hand images. Next, experiments with the artificially degraded images are conducted. Again, in the context of dataset size, this experimental setup is two-fold.  In the first set of experiments, we have considered 150 subjects with the  original and corresponding spoofed samples as a random sub-population from a total of 255 subjects. Next, the other set of experiments considers all the 255 subjects and related fake samples. Unless explicitly mentioned, the experimental description is based on the original and fake sub-population samples consisting of 150 subjects. The results of various experiments with this sub-population are presented in Table 3-6 and pictorially illustrated in Fig.5-7. The experimental outcomes with all 255 subjects are given in Table 7-8 and shown in Fig.8-9.

The objective of our experimental arrangement is delineated as a binary classification to discriminate between the genuine hand images from fake samples using the k-nearest neighbor (k-NN), random forest (RF), and support vector machine (SVM) classifiers. Three (k=3) neighbors are considered using the Euclidean distance for the k-NN classifier. The RF classifier consists of an ensemble of decision trees \textcolor{blue}{\cite{breiman2001random}}. Tree bagging method is used to classify the unknown test samples, and fifty bagged decision trees (RF=1 to 50) have been considered for each experiment. A linear and a radial basis function (RBF) kernels are used in SVM classifiers during testing. 

As only three samples per category are available, two real and original hand images and the corresponding two fake samples per subject are trained, with the remaining one of each category used for testing. In this fashion of choosing the training and testing samples, three different test cases are experimented with one at a time. Finally, the average of three classification results is considered and reported here. The classification errors are measured regarding the false genuine rate (FGR) and false fake rate (FFR).
\begin{itemize}
\item The FGR represents the percentage of spoofed samples are classified as genuine.
\item The FFR denotes the percentage of real images are classified as fake. 
\end{itemize}

The average of these two error rates is the half total error rate, defined as HTER= (FGR+FFR)/2. Before the experiment, IQMs are normalized into [0, 1] scale using the \textit{min-max} rule. 

\begin{table*}[ht]
\centering
  \caption{Different combinations of significant IQMs to minimize error (\%). In each combination, the proposed GMS(\textit{th}=8) remains fixed, and other metrics are concatenated with it one at a time. Considering the following conjugate metrics, the HTERs lie within 0-1.25\%. Significant results for different combinations and classifiers are marked in \textbf{bold} font.  }
  \label{tab:commands}
  \begin{tabular}{l| l| c c c | c c c| c c c | c c c}
    \toprule
  %  \multirow{2}{*}{Dataset} &
    \multicolumn{1}{c} { } & 
      \multicolumn{1}{c} { } &
      \multicolumn{3}{c}{\textbf{k-NN}}&
      \multicolumn{3}{c}{\textbf{RF}} &
      \multicolumn{3}{c}{\textbf{SVM (linear)}}&
      \multicolumn{3}{c}{\textbf{SVM (RBF)}}\\
      Persons & IQMs    & {FFR}	& {FGR} &	{HTER} & {FFR}	& {FGR} &	{HTER} & {FFR}	& {FGR} &	{HTER} & {FFR}	& {FGR} &	{HTER}  \\
      \midrule
\multirow{10}{0.9 cm}{{150 subj.}} & (b) PSNR, (j) GMS & 0 &	0.67 &	0.33 &	0 &	2 &	1 &	0 &	0 &	0 &	0 &	0.67 &	0.33 \\
& (c) SC, (j) GMS &	0 &	0.67&	0.33&	0&	2&	1&	0&	0.67&	0.33&	0&	0.67&	0.33\\

& (g) SSIM, (j) GMS &	0&	2&	1&	0&	1.33&	0.67&	1.33&	1.33&	1.33&	1.33&	1.33&	1.33\\
& (h) ESSIM, (j) GMS &	0 &	2.67 &	1.33 &	0.67 &	2 &	1.33 &	0.67 &	2 &	1.33 &	0.67 &	2 &	1.33\\
& (i) \textbf{WASH}, (j) \textbf{GMS} &	\textbf{0}&	\textbf{2}&	\textbf{1}&\textbf{	0}&\textbf{	0.67}&	\textbf{0.33}&	\textbf{0.67}&	\textbf{1.33}&	\textbf{1}&	\textbf{0.67}&	\textbf{1.33}&	\textbf{1}\\

\cline{2-14}
%\midrule
 & (c) SC, (g) SSIM, (j) GMS & 	0&	0.67&	0.33&	0 &	2 &	1&	0&	0.67&	0.33&	0&	0.67&	0.33\\
 &   (c) SC, (h) ESSIM, (j) GMS &	0&	0&	0&	0.67&	0.67&	0.67&	0&	0&	0&	0&	0&	0\\
 &    (c) \textbf{ SC}, (i) \textbf{WASH}, (j) \textbf{GMS} &	\textbf{0}&	\textbf{0 }&\textbf{	0}&	\textbf{0.67}&\textbf{	0}&	\textbf{0.33}&	\textbf{0}&	\textbf{0}&	\textbf{0}&	\textbf{0}&	\textbf{0}&	\textbf{0}\\
%\midrule
\cline{2-14}
& (c) SC, (h) ESSIM, (i) WASH, (j) GMS & 	0&	0&	0&	0&	2&	    1 &	0&	0&	0&	0&	0&	0\\
 
 \cline{2-14}
 %\midrule
& \textbf{All metrics (a-j) together} & 	\textbf{0}& 	\textbf{0}& 	\textbf{0} & \textbf{0} & 	\textbf{1.33} & 	\textbf{0.67} & 	\textbf{0} & 	\textbf{0} & 	\textbf{0} & 	\textbf{0.67} & 	\textbf{0} & 	\textbf{0.33}
 \\
 \midrule
 \midrule 
 
\multirow{3}{0.9  cm}{{255 subj.}} & (j) \textbf{Proposed GMS (th=8)} & \textbf{1} & \textbf{2}&\textbf{1.5} & \textbf{2} & \textbf{1} &
\textbf{1.5}  & 1.8& 2& 1.9&1.5& 3 & 2.25 \\
 \cline{2-14}
& (i) WASH (j) GMS & 
1 & 3 & 2& 2& 2& 2& 2& 4& 3& 3& 2& 2.5 \\ 
 \cline{2-14}
 & \textbf{All metrics (a-j) together} & \textbf{0} & \textbf{1} & \textbf{0.5} &  \textbf{0} & \textbf{ 1} & \textbf{ 0.5} & \textbf{ 0.75} & \textbf{1.7} & \textbf{1.23}& \textbf{1.2}& \textbf{1.3} & \textbf{1.25} \\
 
    \bottomrule
  \end{tabular}
  \label{table:IQM_combo}
\end{table*}

\begin{table*}
\centering
  \caption{Improvements of proposed GMS over six referred metrics: errors (\%)}
  \label{tab:commands}
  \begin{tabular}{l| c c c | c c c| c c c | c c c}
    \toprule
    %\multirow{2}{*}{Dataset} &
      \multicolumn{1}{c} { } &
      \multicolumn{3}{c}{\textbf{k-NN}}&
      \multicolumn{3}{c}{\textbf{RF}} &
      \multicolumn{3}{c}{\textbf{SVM (linear)}}&
      \multicolumn{3}{c}{\textbf{SVM (RBF)}}\\
      IQM    & {FFR}	& {FGR} &	{HTER} & {FFR}	& {FGR} &	{HTER} & {FFR}	& {FGR} &	{HTER} & {FFR}	& {FGR} &	{HTER}  \\
      \midrule
 (a) MSE	 & 0.67&	2.67&	1.67&	0&	3.33&	1.67&	2&	2&	2&	2&	2&	2\\
(b) PSNR &	  0.67&	2.67&	1.67&	0&	3.33&	1.67&	2&	2&	2&	2&	2&	2\\
(c) SC&	 12.67&	14&	13.33&	11.33&	12	&11.67	&9.33&	7.33&	8.33&	10.67&	4.67&	7.67\\
(d) ED	& 37.33&	38&	37.67&	33.33&	34&	33.67&	56.67&	20&	38.33&	27.33&	34&	30.67\\
(e) CD&	46	&  58.67&	52.33&	59.33&	34&	46.67&	68.67&	21.33&	45&	63.33&	24&	43.67\\
(f) EyD&	40&	  41.33&	40.67&	36&	40&	38&	10&	71.33&	40.67&	24&	67.46&	35.33\\

    \bottomrule
  \end{tabular}
  \label{table:Improvements}
\end{table*}

\begin{figure*}
\centering
% Use the relevant command to insert your figure file.
% For example, with the graphicx package use
\subfigure []{
  \includegraphics[width=0.45\textwidth, keepaspectratio] {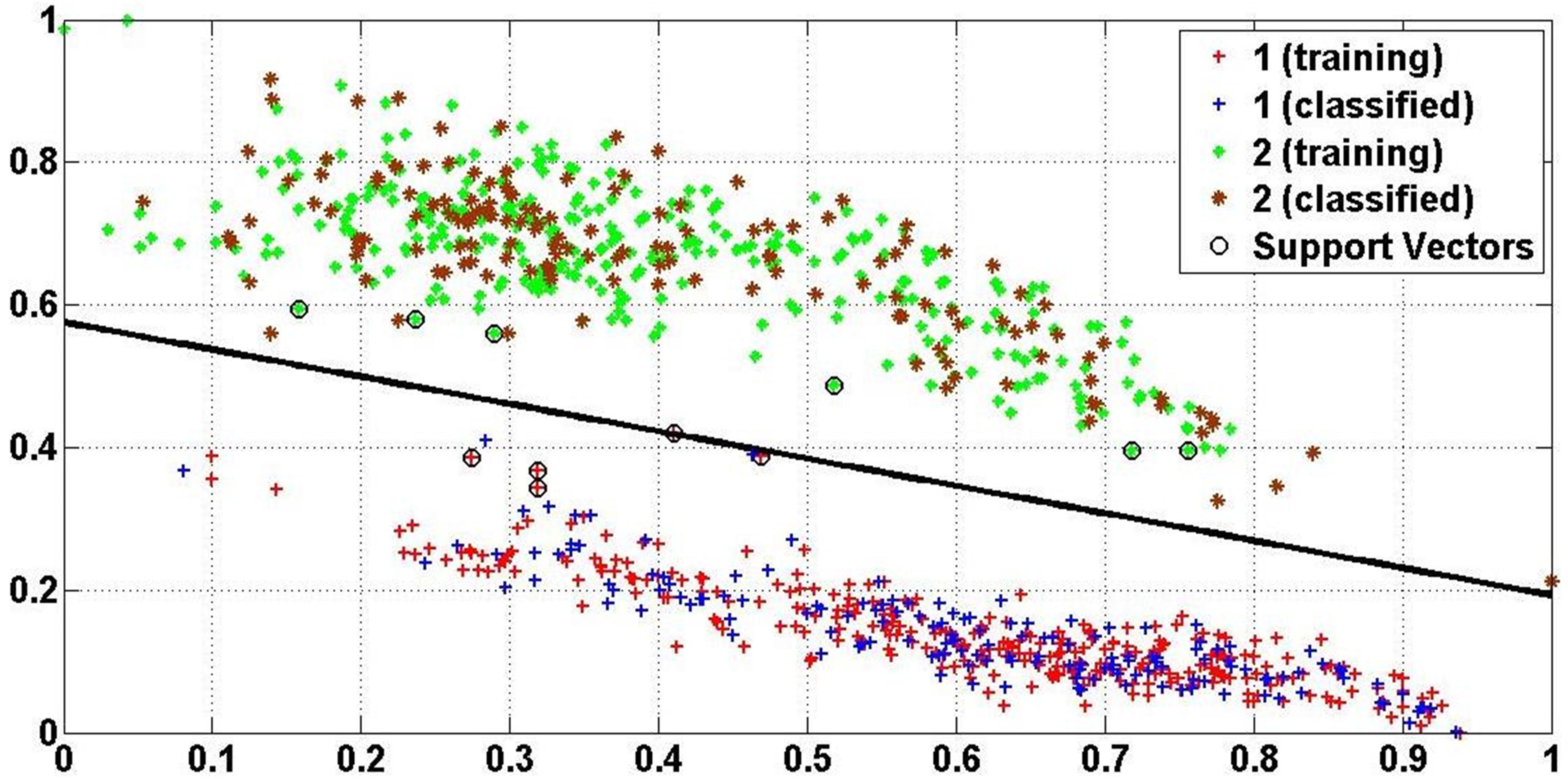}
} 
\subfigure []{
    \includegraphics[width=0.45\textwidth, keepaspectratio] {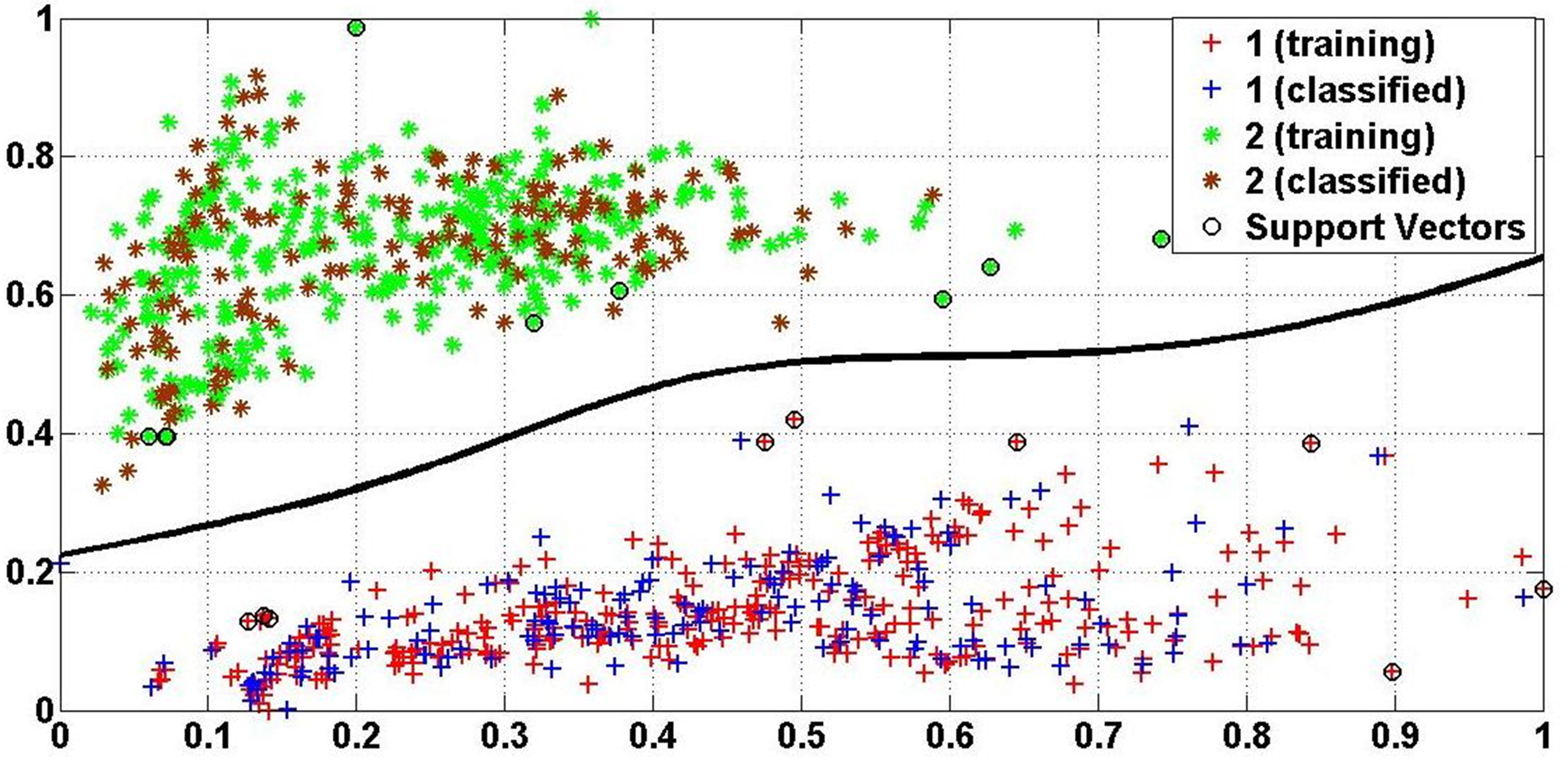}
} 
\hfill    
% figure caption is below the figure
\caption{  Classification with two IQMs: a) PSNR and GMS using SVM with a linear kernel. b) SC and GMS using SVM with an RBF ($\sigma$ =1) kernel. 
}
\label{fig:plot_1}       % Give a unique label
\end{figure*}

\begin{figure*}
\centering
\subfigure []{
  \includegraphics[width=0.45\textwidth, keepaspectratio] {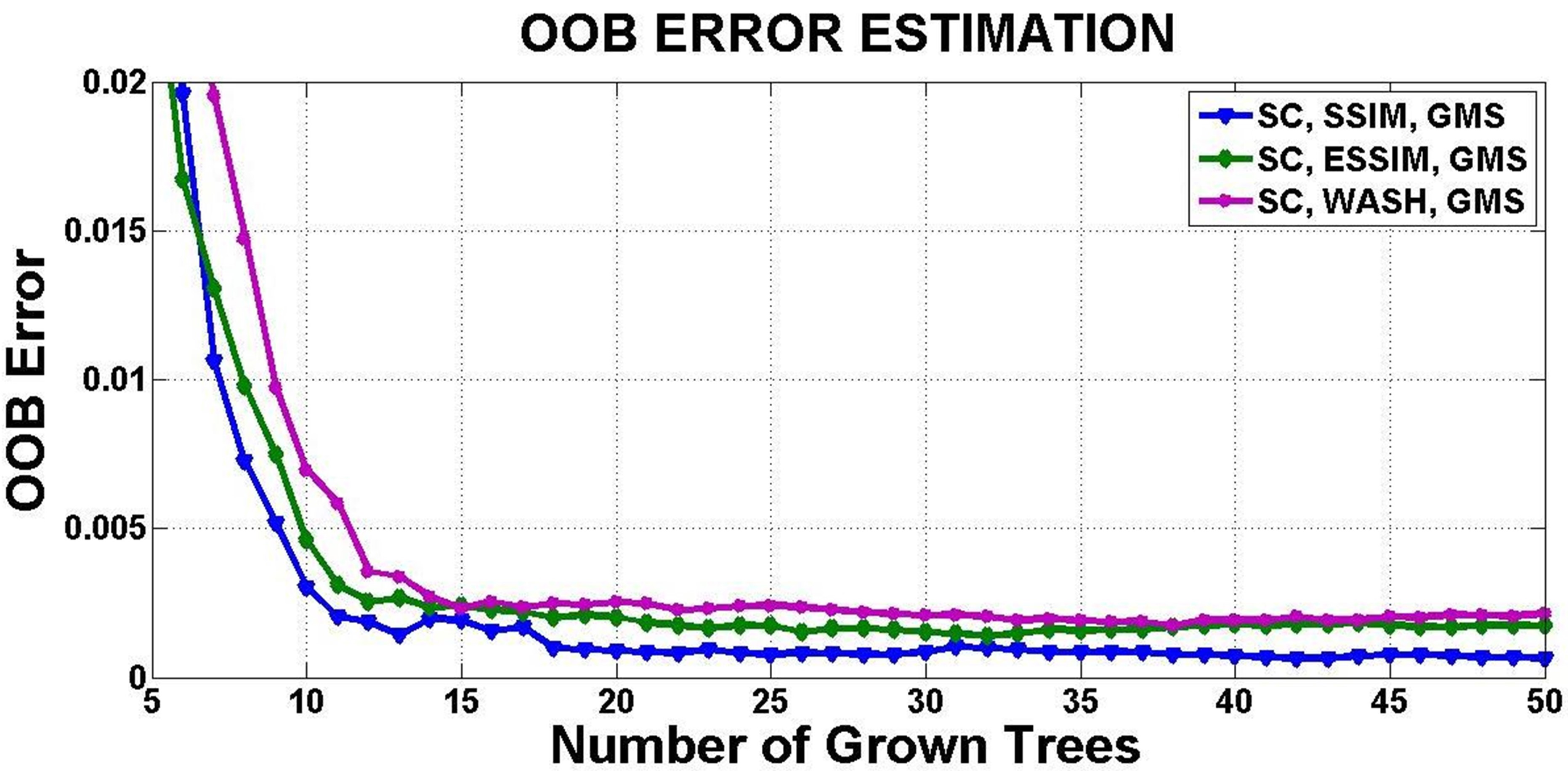}
} 
\subfigure []{
    \includegraphics[width=0.45\textwidth, keepaspectratio] {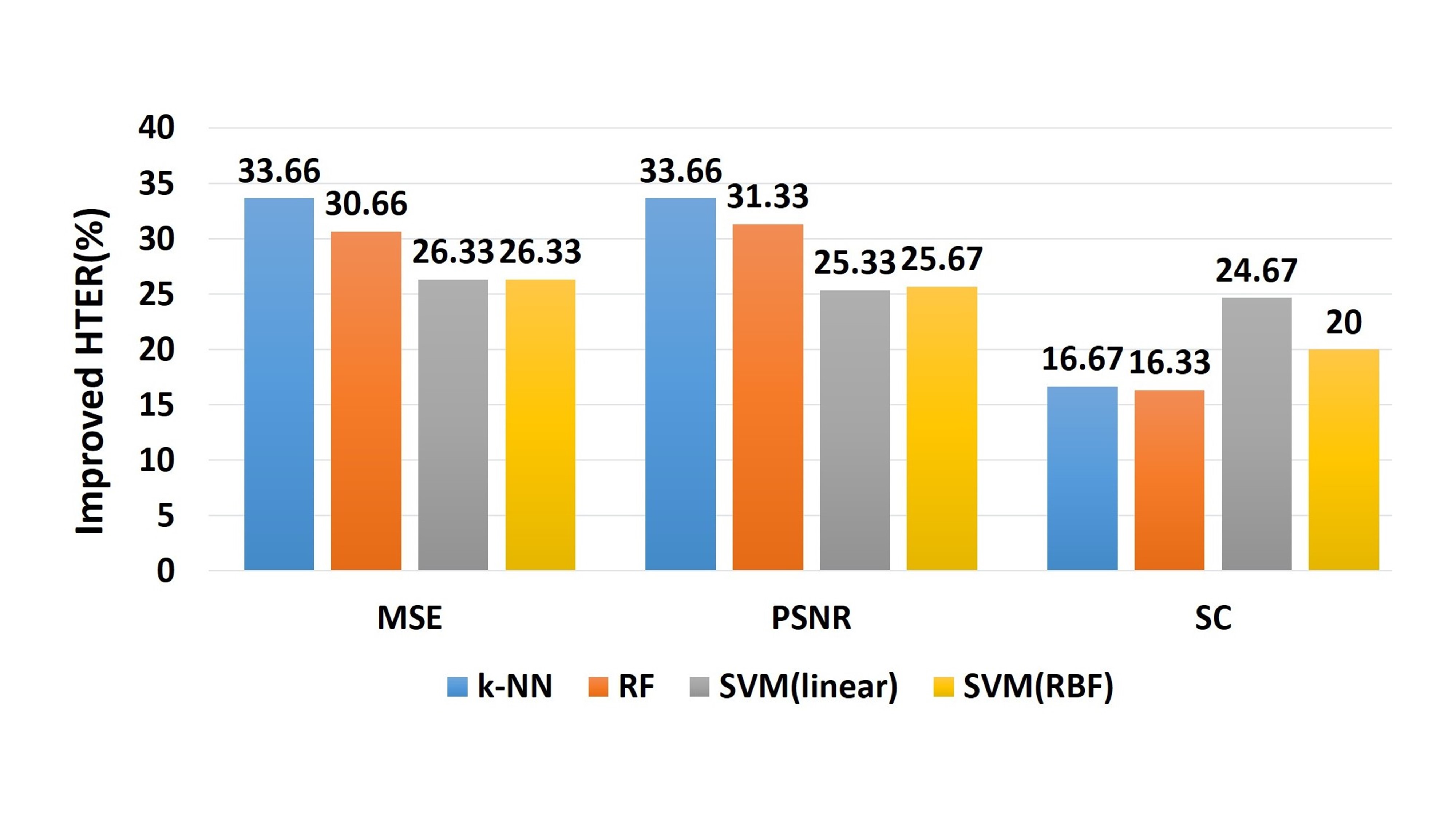}
} 
\hfill    
% figure caption is below the figure
\caption{ a) OOB-error estimation using RF with the SSIM, ESSIM, and WASH chosen one at a time along with the SC and GMS such that only three IQMs are tested in each case. b) The improvements in HTERs due to modification at the preprocessing stage before estimating the MSE, PSNR, and SC quality. 
}
\label{fig:plot_2}       % Give a unique label
\end{figure*}

\begin{figure*}
\centering
\subfigure []{
  \includegraphics[width=0.32\textwidth, keepaspectratio] {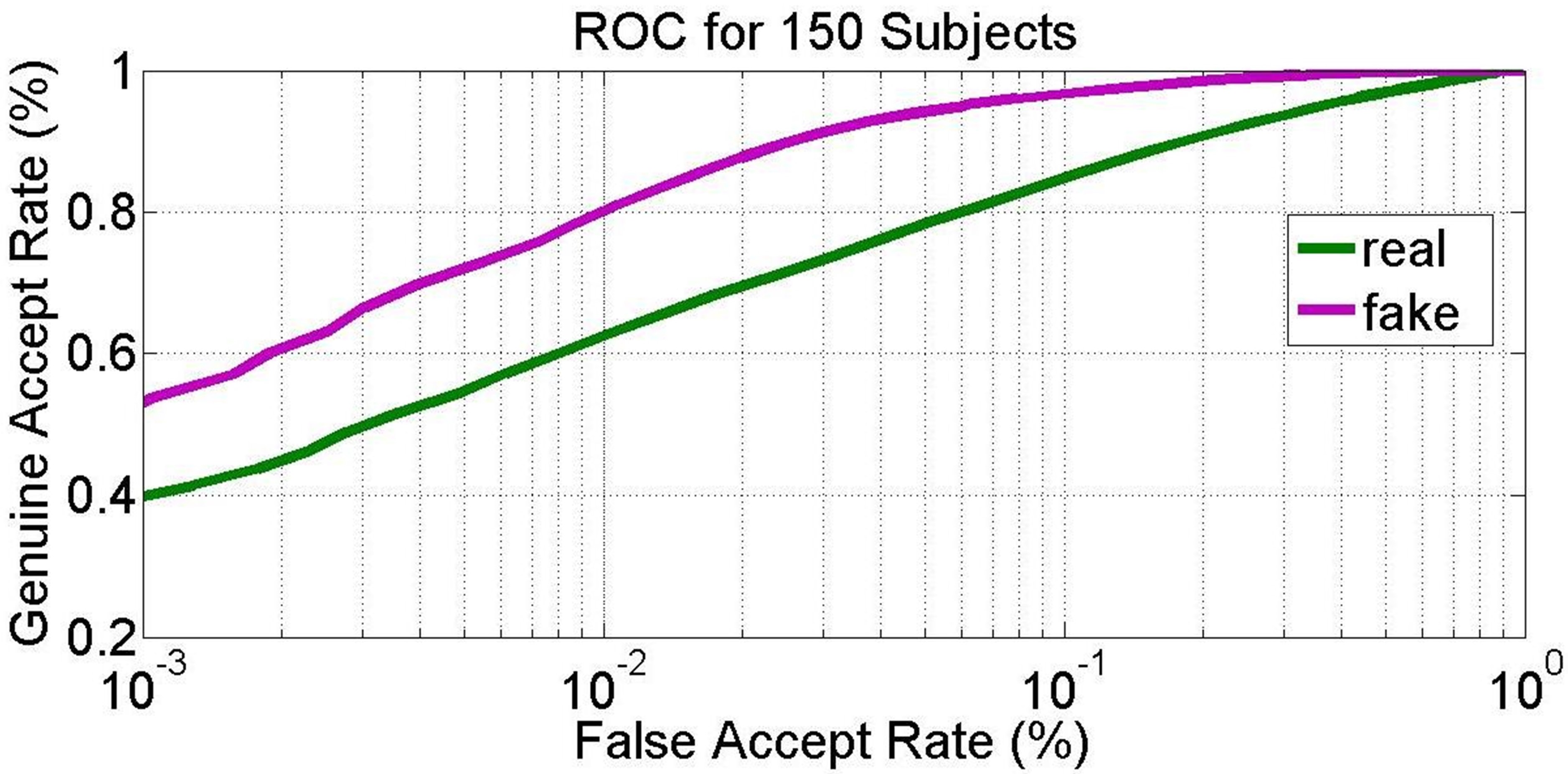}
} 
\subfigure[]{
    \includegraphics[width=0.32\textwidth, keepaspectratio] {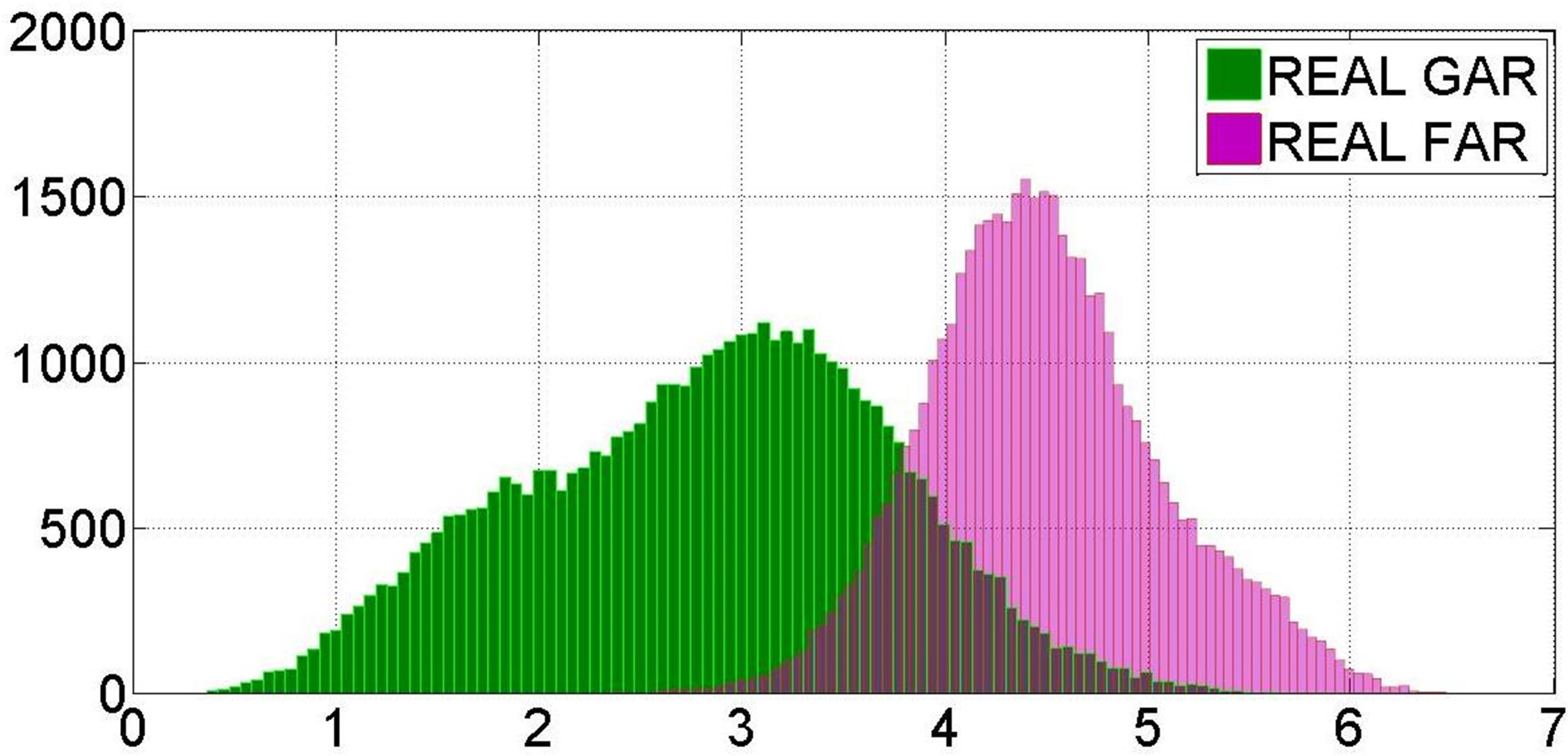}
} 
\subfigure []{
  \includegraphics[width=0.32\textwidth, keepaspectratio] {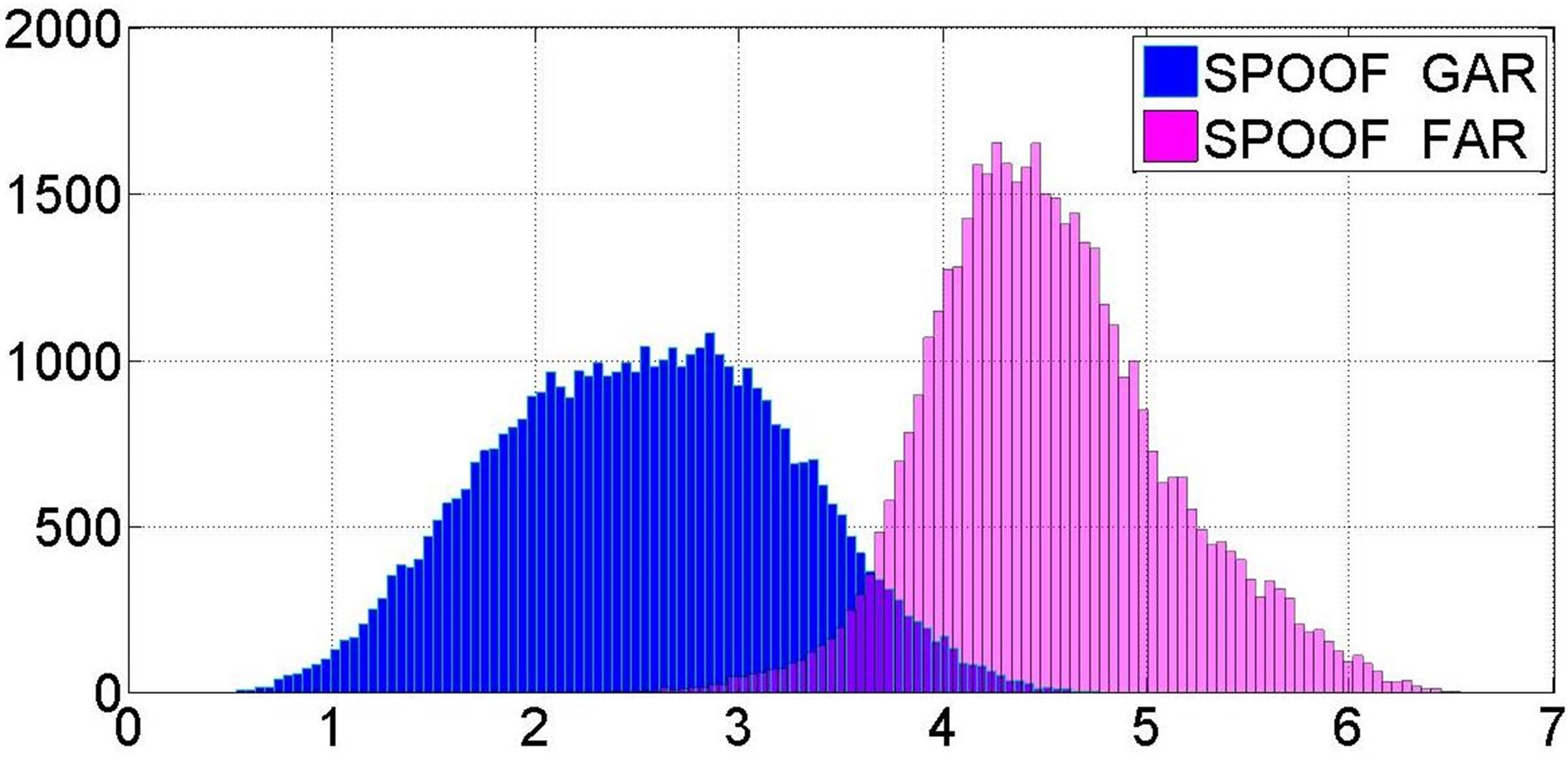}
} 
\hfill    
% figure caption is below the figure
\caption{ a) Receiver Operating Characteristic (ROC) curve illustrates the errors during verification of the original and fake samples.  Histogram distribution of the genuine accept rate (GAR), and the false accept rate (FAR) of (b) real hands and (c) spoofed hands.   
}
\label{fig:plot_3}       % Give a unique label
\end{figure*}

\begin{table*}
\centering
  \caption{Errors (\%) for artificially created fake images from 255 subjects with noise degradation.} 
  \label{tab:commands}
  \begin{tabular}{l| l| c c c | c c c| c c c | c c c}
    \toprule
    %\multirow{2}{*}{Dataset} &
      \multicolumn{1}{c} { } &
      \multicolumn{1}{c} { } &
      \multicolumn{3}{c}{\textbf{k-NN}}&
      \multicolumn{3}{c}{\textbf{RF}} &
      \multicolumn{3}{c}{\textbf{SVM (linear)}}&
      \multicolumn{3}{c}{\textbf{SVM (RBF)}}\\
     Noise & IQM    & {FFR}	& {FGR} &	{HTER} & {FFR}	& {FGR} &	{HTER} & {FFR}	& {FGR} &	{HTER} & {FFR}	& {FGR} &	{HTER}  \\
      \midrule

\multirow{2}{2 cm}{Gaussian Blur }& GMS(th=8)&	 0&	0&	0&	10&	10&	10&	0&	0&	0&	0&	0&	0  	 \\
& All	 &0&	0&	0&	0&	0&	0&	0&	0&	0&	0&	0&	0\\
\midrule

\multirow{2}{2 cm}{Salt and Pepper }& GMS(th=8)	 & 0 &	0&	0&	0&	20&	10&	0&	0&	0&	0&	0&	0 \\
& All	 &  0 &	0&	0&	0&	10&	5&	0&	0&	0&	0&	0&	0	\\

\midrule
\multirow{2}{2 cm}{Speckle}& GMS(th=8)	 &  0 &	0&	0&	10&	20&	15&	0&	0&	0&	0&	0&	0	 \\
& All	 &  0 &	0&	0&	0&	0&	0&	0&	0&	0&	0&	0&	0	\\

    \bottomrule
  \end{tabular}
  \label{table:DL_1}
\end{table*}

\begin{table}
\centering
  \caption{ Error (\%) estimation using MobileNetV2  with 255 subjects}
  \label{tab:commands}
  \begin{tabular}{l| c c c  }
    \toprule
Noise & FFR & FGR & HTER\\
 \midrule
 Natural &	6 &	0 &	3\\
 \midrule
Gaussian Blur  & 6 & 4 & 5  \\
 \midrule
Salt \& pepper &	3 &	2 &	2.5\\
 \midrule
Speckle &	2&	3&	2.5\\

    \bottomrule
  \end{tabular}
  \label{table:DL}
\end{table}

\begin{figure*}
\centering
% Use the relevant command to insert your figure file.
% For example, with the graphicx package use
\subfigure []{
  \includegraphics[width=0.2\textwidth, height= 3.5cm] {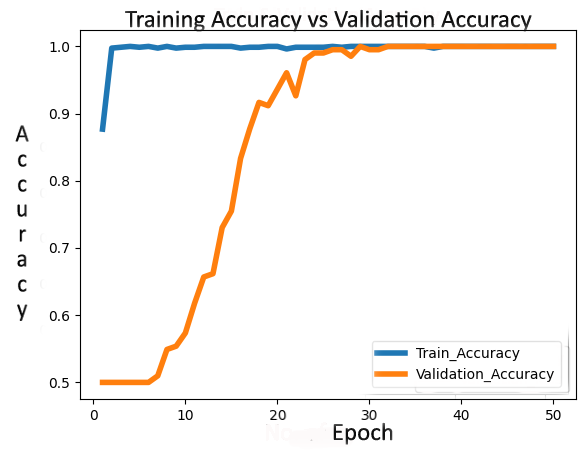}
} 
\subfigure[]{
    \includegraphics[width=0.2\textwidth, height= 3.5cm] {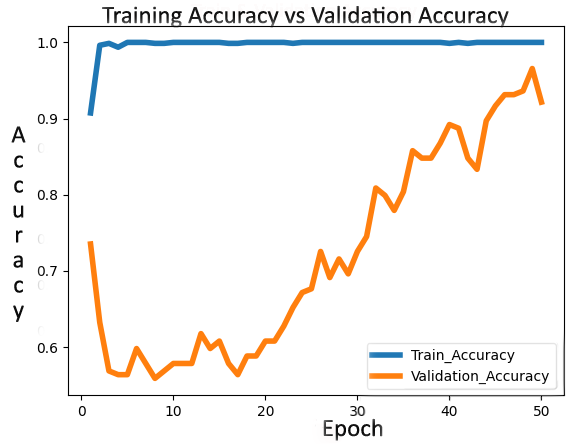}
} 
\subfigure []{
  \includegraphics[width=0.2\textwidth, height= 3.5cm] {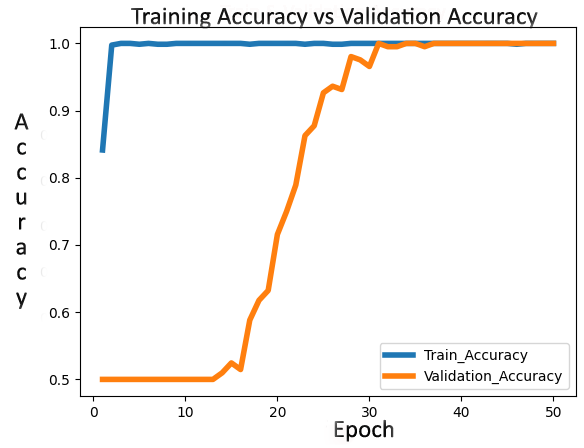}
} 
\subfigure[]{
  \includegraphics[width=0.2\textwidth, height= 3.5cm] {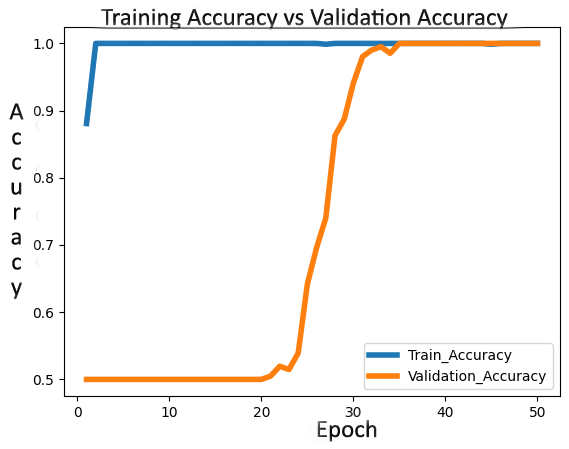}
}
\hfill    
% figure caption is below the figure
\caption{Classification results. a)  accuracy of original and naturally degraded fake images (b) accuracy of original and  degraded fake images using Gaussian blur (c) accuracy of original and  degraded fake images using salt and pepper, and (d) accuracy of original and  degraded fake images using speckle noise. }
\label{fig:CM_1}       % Give a unique label
\end{figure*}

\begin{figure*}
\centering
% Use the relevant command to insert your figure file.
% For example, with the graphicx package use
\subfigure []{
  \includegraphics[width=0.2\textwidth, height= 3.5cm] {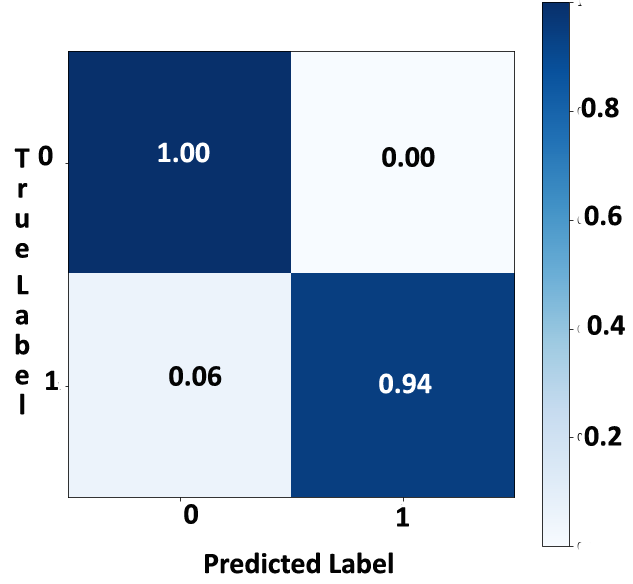}
} 
\subfigure []{
    \includegraphics[width=0.2\textwidth, height= 3.5cm] {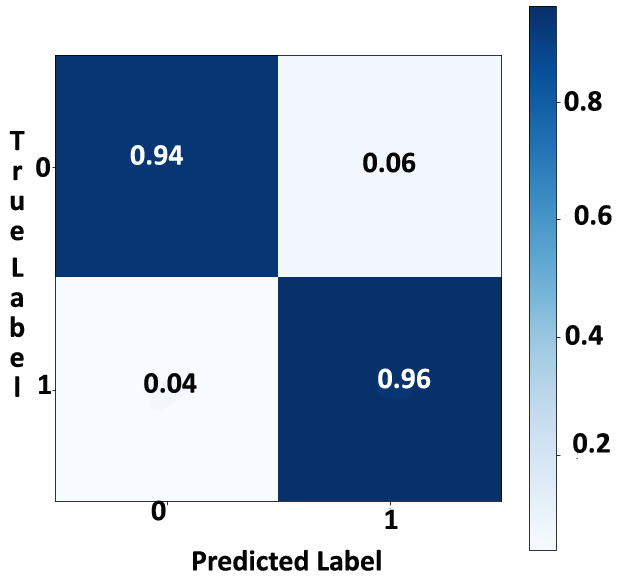}
} 
\subfigure []{
  \includegraphics[width=0.2\textwidth, height= 3.5cm] {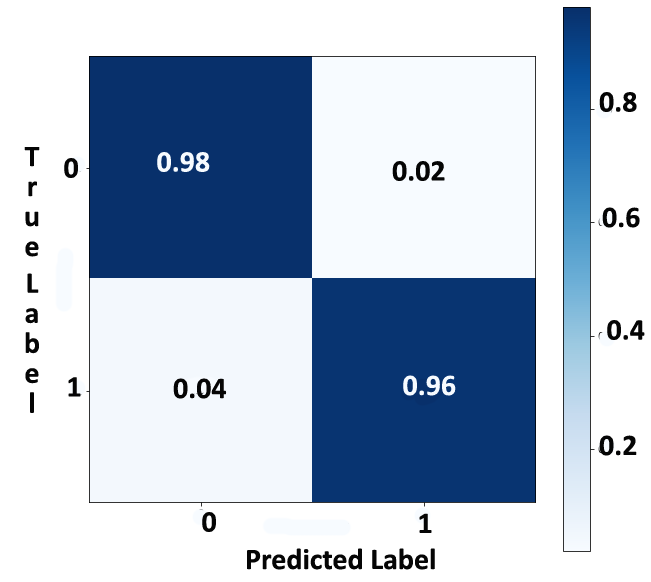}
} 
\subfigure []{
  \includegraphics[width=0.2\textwidth, height= 3.5cm] {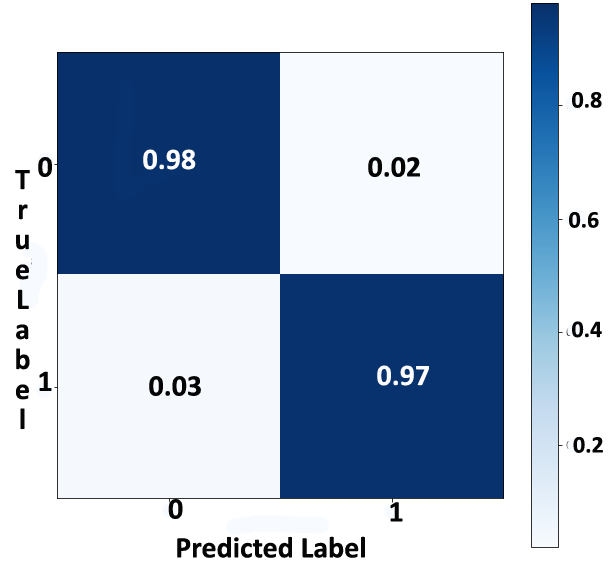}
} 

\hfill    
% figure caption is below the figure
\caption{Confusion matrix. a)  original and naturally degraded fake images (b) original and  degraded fake images using Gaussian blur (c) original and  degraded fake images using salt and pepper, and (d) original and  degraded fake images using speckle noise. }
\label{fig:CM_2}       % Give a unique label
\end{figure*}

Firstly, the quality metrics are evaluated independently to estimate the HTER by the classifiers, as mentioned earlier, and the errors (\%) are provided in Table  \ref{table:HTER_IQM}. The results implicate that SSIM, ESSIM, and GMS (\textit{th}=8) render satisfactory performances. The performance of our proposed GMS is remarkable compared to other existing metrics, particularly, GMS offers comparable errors with the state-of-the-art SSIM and ESSIM metrics. The accuracy of GMS depends on threshold (\textit{th}) values to determine the gradient images. Thus, a good \textit{th} is determined to minimize HTER, and the errors at various thresholds are given in Table \ref{table:GMS}. In this regard, the errors of GMS are determined by two definitions in eq.17 and eq.18. GMS in eq.17 offers better results than eq.18. Thus, the next experiments are carried out with GMS at \textit{th}=8, according to eq.17. GMS standalone cannot render zero-error. Thus, significant metrics are collectively considered with the GMS one at a time and the improvements in results are observed. The HTERs of this test are given in Table \ref{table:IQM_combo}. Mainly, few random subsets containing two, three, and four metrics are tested.  It implies that instead of considering one metric for assessment, we consider two/three metrics as our feature matrix for error estimation. As mentioned earlier, the metrics are normalized to [0, 1] scale prior to experiment. Therefore, the other quality features are concatenated in the feature matrix along the GMS in a simple incremental manner to minimize the error. In this way, we have concatenated different metrics to reduce the errors. Also, all the metrics are evaluated together (Table \ref{table:IQM_combo}) to observe any significant improvement in the results. In this test with 150 subjects, the SVM classifier with both kernels produces comparable errors. Combining all the metrics may not offer  an optimal solution. In this present testing context, only three metrics can be effective for PAD. These three relevant metrics, namely SC, and WASH, with GMS altogether result in 0\% HTERs except RF classifier. An alternative to WASH, the ESSIM can be used as it can offer similar HTER. Thus, it is inferred that only three IQMs are sufficient for anti-spoofing regarding this experimental scenario's hand images of 150 persons.  It is also notable that, all the metrics are all together can not produce the best results. Therefore, for the next experiments with 255 subjects, we have chosen the combination of WASH and GMS metrics to reduce the classification errors because the two metrics produce the minimum HTER for 150 subjects. The HTERs lie within 0.5\% - 1.25\% for different classifiers with all the ten metrics for 255 subjects (Table  \ref{table:IQM_combo}). In this regard, the proposed GMS produces 1.50\% HTER using the k-NN and RF classifiers. The HTER difference between our proposed metric and all the metrics together is at most 1\% using the SVM classifier with RBF kernels. Thus, the performance of GMS (\textit{th}=8) over other existing metric is competitive. This results follow a similar trend as observed with 150 subjects. The HTER of the proposed GMS at \textit{th}=8 is increased to 0.17\% for 255 subjects (Table \ref{table:IQM_combo}) than with 150 subjects (Table  \ref{table:HTER_IQM}) using the k-NN and RF classifiers.

In  another test with the support vectors, only two metrics are tested together, and particularly, GMS is paired with: a) PSNR, and b) SC. These two cases are tested using both the mentioned kernels, and pictorial examples of the support vectors are plotted in Fig.\ref{fig:plot_1}. The GMS and PSNR are tested with a linear kernel SVM (Fig.\ref{fig:plot_1}.a), and GMS with SC is used for classification with the RBF kernel SVM (Fig.\ref{fig:plot_1}.b). Likewise, a combination of four metrics, and all ten metrics are also evaluated. However, no improvement in accuracy has been observed in this experiment. Therefore, only three significant metrics (as specified above) are chosen for our experiments.  

The out-of-bag (OOB) errors are estimated by RF classifier with a subset of three good features during this experiment. It computes the average of cumulative  wrong classification probability for OOB observations in the bootstrap dataset. The minimum OOB errors are: a) SC, SSIM, and GMS: $6.41 \times 10^{-4} $; b) SC, ESSIM, and GMS: $1.4 \times 10^{-3} $; and c) SC, WASH, and GMS: $1.7 \times 10^{-3} $. These OOB errors are estimated with 1 to 50 decisions trees. A graphical comparison of OOB errors with these combinations of metrics is shown in Fig.\ref{fig:plot_2}.a. It is clear that SC, SSIM, and GMS collectively performs the best for correct classification. 

In another experiment, we have introduced a slight modification during the preprocessing stage before computing the defined metrics. In this test, six IQMs particularly, the MSE, PSNR, SC, ED, CD, and EyD are computed with this new preprocessing step.  Firstly, the original hand image is considered, and its gradient image is computed at \textit{th}=8 using eq.13-14. Next, the specified IQMs are extracted. Similarly, the fake images are smoothed with the same Gaussian filter, and the gradient image is calculated in the same manner as applied for original hands. The results are given in Table \ref{table:Improvements}. This modification at the preprocessing on images for these metrics also produces remarkable results than the errors reported in Table \ref{table:HTER_IQM}. In this test, reduced errors(\%) are attained (Table \ref{table:Improvements}), and significant gains in HTERs are shown in Fig.\ref{fig:plot_2}.b. However, very small gains in HTERs are obtained using RF for ED (1\%), and CD (2\%); while no improvement has been observed for EyD by any classifier. Hence, the present preprocessing is also effective to compute HTERs for existing IQMs. It justifies that our  thresholded gradient computation can also be useful at the preprocessing stage  to compute other metrics.

In the next experiment, we have considered the original and fake hand samples of all 255 subjects. Similar to the aforesaid  experimental strategy, three classification tests are conducted. Specifically, these tests discriminates between the real hand images and all three types of fake hand images which are artificially degraded with the Gaussian blur, salt and pepper noise, and speckle noise. We have used  765 original (real) and 765 fake (total 1530) images for each of  the classification tasks.
Here, we consider our proposed GMS metric at \textit{th}=8 and all the metrics together for classification. The results are given in Table \ref{table:DL_1}. The performance (HTER) of the proposed GMS metric is excellent in discriminating artificially degraded fake and original samples using the SVM classifier with both types of kernel. Therefore, the proposed metric performs well for spoofing detection using fake hand samples with noise degradation.

Next, the verification of real and spoofed templates is experimented regarding a decision threshold $t_v$, defined as:
\begin{equation}
t_v=  \sqrt{ \mathop{\sum^{W}_{f=1}} \frac{(U_f -V_f)^2}{\sigma ^2} }
\end{equation}
where an IQM is denoted with $\textit{f}$ trained samples of a subject is denoted by $\textit{U}$; an unknown IQM of a claimer is represented with $\textit{V}$, and $\sigma$ is the standard deviation of quality determinative feature $\textit{f}$. The verification performance is calculated regarding:
\begin{itemize}
\item False Accept Rate (FAR): a spoofed matching score is lower than the threshold $t_v$.
\item False Reject Rate (FRR): a legitimate matching score is more than the threshold $t_v$.
\item Genuine Accept Rate (GAR) = 1 - FAR 
\end{itemize}

The GAR and FAR depend on $t_v$, and are plotted in a Receiver Operating Characteristic (ROC) curve \textcolor{blue}{\cite{gamassi2005quality}}, in Fig.\ref{fig:plot_3}.a. Alternatively, the GAR and FAR of genuine and spoofed samples are considered separately through the histograms with 100 bins for representing the quality template matching distances. The histogram distributions are plotted in Fig.\ref{fig:plot_3}.b-c. 

\subsubsection{Deep learning experiment}
\vspace{-0.3 cm}
Deep learning (DL) models are widely used for solving image recognition problems in recent years. We have experimented with our spoofing detection approach on our fake datasets using  the DL method. As the conventional heavyweight CNNs (e.g. VGG-19, ResNet-50) follow deeper and complex network architecture, stacking multiple convolution layers, therefore, are time consuming. To cope with the time-constraint, we have used the MobileNetV2 model \textcolor{blue}{\cite {sandler2018mobilenetv2}}. It is a lightweight model as compared to other deep models, as stated above. It can also  be used in the android environment for classification task like ours. 

For our experiment, a binary classification model is created leveraging the high-level feature maps of the MobileNetV2 base network using the Tensorflow and Keras framework. 
Similar to the aforesaid  experimental strategy, four classification tests are conducted with all subjects. In these test cases, the dataset is split with 8:1:1 ratio, where 80\% of total images are used for training, and 10\%  of images are for validation, and remaining 10\% for testing purposes. We have applied standard data augmentation techniques (i.e., horizontal flip, zoom (20\%), shearing (20\%), and rotation 90 degrees) with random variations. The original color hand image with 382 $\times$ 585 pixels is resized to 224 $\times$ 224 pixels. 
We have used a simple transfer learning approach. We have adapted pre-trained Imagenet weight in base CNN for faster convergence.  The Stochastic Gradient Descent (SGD) optimizer with a momentum of 0.90 is applied to reduce the categorical cross-entropy loss function, and the learning rate is 0.0001.  The model is trained for a total of 100 epochs in two-steps with a mini-batch size 8. First, our model is trained for 50 epochs, resulting in an uneven training and validation performance. Next, the model is trained further using the saved weights of step-1 for 50 more epochs with same batch size. Fig.\ref{fig:CM_1} clearly shows the training accuracy vs. validation accuracy for each type of fake samples for step-2.
It is evident that the training behaviours of these tests' are similar, and a little variation is observed in the validation accuracy due to different fake samples with noise degradation. The trained model is tested with the remaining unseen 10\% images of each experiment's total dataset. The classification accuracy(\%) is given in Table \ref{table:DL}. The confusion matrix of each experiment is shown in Fig.\ref{fig:CM_2}. The deep network effectively distinguishes the real and fake hand images. From all the experiments, it is clear that the natural degraded images are challenging to detect than  artificial degradation with salt and pepper noise and speckle noise. However, Gaussian noise is more challenging for correct spoofing detection.

In \textcolor{blue}{\cite{bartuzi2018thermal}}, thermal features of the hand are computed using a CNN architecture. It applies a score fusion of the RGB and thermal image matching modules to reach a final decision. According to our study, no comparable work is available for anti-spoofing on hand images using a software-based method such as the IQMs, which is explored in this proposed work. However, there exist a few hardware-based liveness detection methods using hand images. Therefore, there is no scope for performance assessment than other available similar works on the same hand image dataset. 

There is a further scope to experiment with more challenging fake samples for PAD. Spoofing detection using real-time hand images will be more challenging in a realistic situation. Presently, our method is evaluated on a small dataset with only 255 subjects. Therefore, the fake hand dataset will be increased to a larger population and should be made publicly available for further exploration in this direction. Also, different sensors can be used for fake image acquisition, and can be tested with our proposed method for better generalization. It will be interesting to explore further real-time fake samples in a mobile/android based environment, which is currently a limitation of our approach. In future, anti-spoofing detection on hand biometrics can be investigated with more challenging datasets using the deep learning methods. Lastly, new quality metrics can be devised for further improvement in the context of hand biometric spoofing detection. 

\vspace{-0.3 cm}
\section{Conclusion }
In this paper, a method for presentation attack detection (PAD) using the visual quality assessment is presented to defend illegitimate attempts for a hand biometric system. The defined quality metrics are very significant for anti-spoofing on real hand images. A new gradient magnitude-based FR-IQM is described that renders satisfactory classification results. The proposed GMS defends electronic display based spoofing attack (i.e. natural degradation)  on hand images with at most 3\% HTER using traditional hand-crafted and deep learning experimental settings. The fake samples are discriminated from original samples using deep networks along with the conventional classification methods. This method can be useful for software-based fake sample detection at an earlier stage of a deployed hand biometric system in a cost-efficient manner. However, no experiment for hand biometric authentication has been reported here, which is further scope to enhance this work such that the zero-effort imposters can also be detected successfully. We plan to investigate further for necessary counter-measurements to deter spoofing-attacks on hand biometrics.

% BibTeX users please use one of
\bibliographystyle{spbasic}      % basic style, author-year citations
%\bibliographystyle{spmpsci}      % mathematics and physical sciences
%\bibliographystyle{spphys}       % APS-like style for physics
%\bibliography{}   % name your BibTeX data base

\bibliography{Ref.bib}

\begin{thebibliography}{58}
\providecommand{\natexlab}[1]{#1}
\providecommand{\url}[1]{{#1}}
\providecommand{\urlprefix}{URL }
\expandafter\ifx\csname urlstyle\endcsname\relax
  \providecommand{\doi}[1]{DOI~\discretionary{}{}{}#1}\else
  \providecommand{\doi}{DOI~\discretionary{}{}{}\begingroup
  \urlstyle{rm}\Url}\fi
\providecommand{\eprint}[2][]{\url{#2}}

\bibitem[{Banitalebi-Dehkordi et~al.(2019)Banitalebi-Dehkordi, Khademi,
  Ebrahimi-Moghadam, and Hadizadeh}]{banitalebi2019image}
Banitalebi-Dehkordi M, Khademi M, Ebrahimi-Moghadam A, Hadizadeh H (2019) An
  image quality assessment algorithm based on saliency and sparsity. Multimedia
  Tools and Applications 78(9):11507--11526

\bibitem[{Bapat and Kanhangad(2017)}]{bapat2017segmentation}
Bapat A, Kanhangad V (2017) Segmentation of hand from cluttered backgrounds for
  hand geometry biometrics. In: 2017 IEEE Region 10 Symposium (TENSYMP), IEEE,
  pp 1--4

\bibitem[{Barra et~al.(2019)Barra, De~Marsico, Nappi, Narducci, and
  Riccio}]{barra2019hand}
Barra S, De~Marsico M, Nappi M, Narducci F, Riccio D (2019) A hand-based
  biometric system in visible light for mobile environments. Information
  Sciences 479:472--485

\bibitem[{Bartuzi and Trokielewicz(2018)}]{bartuzi2018thermal}
Bartuzi E, Trokielewicz M (2018) Thermal features for presentation attack
  detection in hand biometrics. In: 2018 IEEE 9th International Conference on
  Biometrics Theory, Applications and Systems (BTAS), IEEE, pp 1--6

\bibitem[{{Bera} and {Bhattacharjee}(2020)}]{Bera8032481}
{Bera} A, {Bhattacharjee} D (2020) Human identification using selected features
  from finger geometric profiles. IEEE Transactions on Systems, Man, and
  Cybernetics: Systems 50(3):747--761

\bibitem[{Bera et~al.(2014)Bera, Bhattacharjee, and Nasipuri}]{bera2014hand}
Bera A, Bhattacharjee D, Nasipuri M (2014) Hand biometrics in digital
  forensics. In: Computational Intelligence in Digital Forensics: Forensic
  Investigation and Applications, Springer, pp 145--163

\bibitem[{Bera et~al.(2015)Bera, Bhattacharjee, and Nasipuri}]{bera2015fusion}
Bera A, Bhattacharjee D, Nasipuri M (2015) Fusion-based hand geometry
  recognition using dempster--shafer theory. International Journal of Pattern
  Recognition and Artificial Intelligence 29(05):1556005

\bibitem[{Bera et~al.(2017)Bera, Bhattacharjee, and Nasipuri}]{bera2017finger}
Bera A, Bhattacharjee D, Nasipuri M (2017) Finger contour profile based hand
  biometric recognition. Multimedia Tools and Applications 76(20):21451--21479

\bibitem[{Bera et~al.(2019)Bera, Bhattacharjee, and Nasipuri}]{bera2019finger}
Bera A, Bhattacharjee D, Nasipuri M (2019) Finger biometric recognition with
  feature selection. The Biometric Computing: Recognition and Registration p~87

\bibitem[{Bhilare et~al.(2018)Bhilare, Kanhangad, and
  Chaudhari}]{bhilare2018study}
Bhilare S, Kanhangad V, Chaudhari N (2018) A study on vulnerability and
  presentation attack detection in palmprint verification system. Pattern
  Analysis and Applications 21(3):769--782

\bibitem[{Biggio et~al.(2016)Biggio, Fumera, Marcialis, and
  Roli}]{biggio2016statistical}
Biggio B, Fumera G, Marcialis GL, Roli F (2016) Statistical meta-analysis of
  presentation attacks for secure multibiometric systems. IEEE transactions on
  pattern analysis and machine intelligence 39(3):561--575

\bibitem[{Bondzulic et~al.(2018)Bondzulic, Petrovic, Andric, and
  Pavlovic}]{bondzulic2018gradient}
Bondzulic B, Petrovic V, Andric M, Pavlovic B (2018) Gradient-based image
  quality assessment. Acta Polytechnica Hungarica 15(4)

\bibitem[{Bong and Khoo(2014)}]{bong2014blind}
Bong DBL, Khoo BE (2014) Blind image blur assessment by using valid reblur
  range and histogram shape difference. Signal Processing: Image Communication
  29(6):699--710

\bibitem[{Breiman(2001)}]{breiman2001random}
Breiman L (2001) Random forests. Machine learning 45(1):5--32

\bibitem[{Chatterjee et~al.(2018)Chatterjee, Singh, Bhatia, and
  Prakash}]{chatterjee2018low}
Chatterjee A, Singh P, Bhatia V, Prakash S (2018) A low-cost optical sensor for
  secured antispoof touchless palm print biometry. IEEE sensors letters
  2(2):1--4

\bibitem[{Chen et~al.(2005)Chen, Valizadegan, Jackson, Soltysiak, and
  Jain}]{chen2005fake}
Chen H, Valizadegan H, Jackson C, Soltysiak S, Jain AK (2005) Fake hands:
  spoofing hand geometry systems. Biometric Consortium

\bibitem[{Chen et~al.(2016)Chen, Wang, Yang, and He}]{chen2016finger}
Chen L, Wang J, Yang S, He H (2016) A finger vein image-based personal
  identification system with self-adaptive illuminance control. IEEE
  Transactions on Instrumentation and Measurement 66(2):294--304

\bibitem[{Chingovska et~al.(2014)Chingovska, Dos~Anjos, and
  Marcel}]{chingovska2014biometrics}
Chingovska I, Dos~Anjos AR, Marcel S (2014) Biometrics evaluation under
  spoofing attacks. IEEE transactions on Information Forensics and Security
  9(12):2264--2276

\bibitem[{Chugh et~al.(2018)Chugh, Cao, and Jain}]{chugh2018fingerprint}
Chugh T, Cao K, Jain AK (2018) Fingerprint spoof buster: Use of
  minutiae-centered patches. IEEE Transactions on Information Forensics and
  Security 13(9):2190--2202

\bibitem[{Czajka and Bulwan(2013)}]{czajka2013biometric}
Czajka A, Bulwan P (2013) Biometric verification based on hand thermal images.
  In: 2013 International Conference on Biometrics (ICB), IEEE, pp 1--6

\bibitem[{Doi and Yamanaka(2005)}]{doi2005discrete}
Doi J, Yamanaka M (2005) Discrete finger and palmar feature extraction for
  personal authentication. IEEE transactions on instrumentation and measurement
  54(6):2213--2219

\bibitem[{Dutagaci et~al.(2008)Dutagaci, Sankur, and
  Y{\"o}r{\"u}k}]{dutagaci2008comparative}
Dutagaci H, Sankur B, Y{\"o}r{\"u}k E (2008) Comparative analysis of global
  hand appearance-based person recognition. Journal of electronic imaging
  17(1):011018

\bibitem[{Farmanbar and Toygar(2017)}]{farmanbar2017spoof}
Farmanbar M, Toygar {\"O} (2017) Spoof detection on face and palmprint
  biometrics. Signal, Image and Video Processing 11(7):1253--1260

\bibitem[{Faundez-Zanuy et~al.(2014)Faundez-Zanuy, Mekyska, and
  Font-Aragon{\`e}s}]{faundez2014new}
Faundez-Zanuy M, Mekyska J, Font-Aragon{\`e}s X (2014) A new hand image
  database simultaneously acquired in visible, near-infrared and thermal
  spectrums. Cognitive Computation 6(2):230--240

\bibitem[{Ferrer et~al.(2014)Ferrer, Morales, and
  D{\'\i}az}]{ferrer2014approach}
Ferrer MA, Morales A, D{\'\i}az A (2014) An approach to swir hyperspectral hand
  biometrics. Information Sciences 268:3--19

\bibitem[{Fourati et~al.(2020)Fourati, Elloumi, and
  Chetouani}]{fourati2020anti}
Fourati E, Elloumi W, Chetouani A (2020) Anti-spoofing in face
  recognition-based biometric authentication using image quality assessment.
  Multimedia Tools and Applications 79(1-2):865--889

\bibitem[{Galbally et~al.(2014)Galbally, Marcel, and
  Fierrez}]{galbally2013image}
Galbally J, Marcel S, Fierrez J (2014) Image quality assessment for fake
  biometric detection: Application to iris, fingerprint, and face recognition.
  IEEE transactions on image processing 23(2):710--724

\bibitem[{Gamassi et~al.(2005)Gamassi, Lazzaroni, Misino, Piuri, Sana, and
  Scotti}]{gamassi2005quality}
Gamassi M, Lazzaroni M, Misino M, Piuri V, Sana D, Scotti F (2005) Quality
  assessment of biometric systems: a comprehensive perspective based on
  accuracy and performance measurement. IEEE Transactions on Instrumentation
  and Measurement 54(4):1489--1496

\bibitem[{Gao et~al.(2018)Gao, Miao, Yang, and Ma}]{gao2018image}
Gao H, Miao Q, Yang J, Ma Z (2018) Image quality assessment using image
  description in information theory. IEEE Access 6:47181--47188

\bibitem[{Guan et~al.(2017)Guan, Yi, Zeng, Cham, and Wang}]{guan2017visual}
Guan J, Yi S, Zeng X, Cham WK, Wang X (2017) Visual importance and distortion
  guided deep image quality assessment framework. IEEE Transactions on
  Multimedia 19(11):2505--2520

\bibitem[{Harvey et~al.(2018)Harvey, Campbell, and
  Adler}]{harvey2018characterization}
Harvey J, Campbell J, Adler A (2018) Characterization of biometric template
  aging in a multiyear, multivendor longitudinal fingerprint matching study.
  IEEE Transactions on Instrumentation and Measurement 68(4):1071--1079

\bibitem[{Jaswal et~al.(2019)Jaswal, Kaul, and Nath}]{jaswal2019multimodal}
Jaswal G, Kaul A, Nath R (2019) Multimodal biometric authentication system
  using hand shape, palm print, and hand geometry. In: Computational
  Intelligence: Theories, Applications and Future Directions-Volume II,
  Springer, pp 557--570

\bibitem[{Jia et~al.(2020)Jia, Guo, Xu, and Wang}]{jia2020face}
Jia S, Guo G, Xu Z, Wang Q (2020) Face presentation attack detection in mobile
  scenarios: A comprehensive evaluation. Image and Vision Computing 93:103826

\bibitem[{Klonowski et~al.(2018)Klonowski, Plata, and Syga}]{klonowski2018user}
Klonowski M, Plata M, Syga P (2018) User authorization based on hand geometry
  without special equipment. Pattern Recognition 73:189--201

\bibitem[{Korshunov and Marcel(2017)}]{korshunov2017impact}
Korshunov P, Marcel S (2017) Impact of score fusion on voice biometrics and
  presentation attack detection in cross-database evaluations. IEEE Journal of
  Selected Topics in Signal Processing 11(4):695--705

\bibitem[{Liu et~al.(2011)Liu, Lin, and Narwaria}]{liu2011image}
Liu A, Lin W, Narwaria M (2011) Image quality assessment based on gradient
  similarity. IEEE Transactions on Image Processing 21(4):1500--1512

\bibitem[{Martini et~al.(2012)Martini, Hewage, and
  Villarini}]{martini2012image}
Martini MG, Hewage CT, Villarini B (2012) Image quality assessment based on
  edge preservation. Signal Processing: Image Communication 27(8):875--882

\bibitem[{Nogueira et~al.(2016)Nogueira, de~Alencar~Lotufo, and
  Machado}]{nogueira2016fingerprint}
Nogueira RF, de~Alencar~Lotufo R, Machado RC (2016) Fingerprint liveness
  detection using convolutional neural networks. IEEE transactions on
  information forensics and security 11(6):1206--1213

\bibitem[{Patil et~al.(2016)Patil, Bhilare, and Kanhangad}]{patil2016assessing}
Patil I, Bhilare S, Kanhangad V (2016) Assessing vulnerability of dorsal
  hand-vein verification system to spoofing attacks using smartphone camera.
  In: 2016 IEEE International Conference on Identity, Security and Behavior
  Analysis (ISBA), IEEE, pp 1--6

\bibitem[{Pinto et~al.(2020)Pinto, Goldenstein, Ferreira, Carvalho, Pedrini,
  and Rocha}]{pinto2020leveraging}
Pinto A, Goldenstein S, Ferreira A, Carvalho T, Pedrini H, Rocha A (2020)
  Leveraging shape, reflectance and albedo from shading for face presentation
  attack detection. IEEE Transactions on Information Forensics and Security
  15:3347--3358

\bibitem[{Qiu et~al.(2017)Qiu, Kang, Tian, Jia, and Huang}]{qiu2017finger}
Qiu X, Kang W, Tian S, Jia W, Huang Z (2017) Finger vein presentation attack
  detection using total variation decomposition. IEEE Transactions on
  Information Forensics and Security 13(2):465--477

\bibitem[{Raghavendra and Busch(2015)}]{raghavendra2015robust}
Raghavendra R, Busch C (2015) Robust scheme for iris presentation attack
  detection using multiscale binarized statistical image features. IEEE
  Transactions on Information Forensics and Security 10(4):703--715

\bibitem[{Rahul and Tiwari(2019)}]{rahul2019fqi}
Rahul K, Tiwari AK (2019) Fqi: feature-based reduced-reference image quality
  assessment method for screen content images. IET Image Processing
  13(7):1170--1180

\bibitem[{Rathgeb et~al.(2020)Rathgeb, Drozdowski, Fischer, and
  Busch}]{rathgeb2020vulnerability}
Rathgeb C, Drozdowski P, Fischer D, Busch C (2020) Vulnerability assessment and
  detection of makeup presentation attacks. In: 2020 8th International Workshop
  on Biometrics and Forensics (IWBF), IEEE, pp 1--6

\bibitem[{Reenu et~al.(2013)Reenu, David, Raj, and Nair}]{reenu2013wavelet}
Reenu M, David D, Raj SA, Nair MS (2013) Wavelet based sharp features (wash):
  An image quality assessment metric based on hvs. In: 2013 2nd International
  Conference on Advanced Computing, Networking and Security, IEEE, pp 79--83

\bibitem[{Sajjad et~al.(2019)Sajjad, Khan, Hussain, Muhammad, Sangaiah,
  Castiglione, Esposito, and Baik}]{sajjad2019cnn}
Sajjad M, Khan S, Hussain T, Muhammad K, Sangaiah AK, Castiglione A, Esposito
  C, Baik SW (2019) Cnn-based anti-spoofing two-tier multi-factor
  authentication system. Pattern Recognition Letters 126:123--131

\bibitem[{Sandler et~al.(2018)Sandler, Howard, Zhu, Zhmoginov, and
  Chen}]{sandler2018mobilenetv2}
Sandler M, Howard A, Zhu M, Zhmoginov A, Chen LC (2018) Mobilenetv2: Inverted
  residuals and linear bottlenecks. In: Proceedings of the IEEE conference on
  computer vision and pattern recognition, pp 4510--4520

\bibitem[{Sellahewa and Jassim(2010)}]{sellahewa2010image}
Sellahewa H, Jassim SA (2010) Image-quality-based adaptive face recognition.
  IEEE Transactions on Instrumentation and measurement 59(4):805--813

\bibitem[{Sun et~al.(2018)Sun, Liao, Xue, and Zhou}]{sun2018spsim}
Sun W, Liao Q, Xue JH, Zhou F (2018) Spsim: A superpixel-based similarity index
  for full-reference image quality assessment. IEEE Transactions on Image
  Processing 27(9):4232--4244

\bibitem[{Tolosana et~al.(2019)Tolosana, Gomez-Barrero, Busch, and
  Ortega-Garcia}]{tolosana2019biometric}
Tolosana R, Gomez-Barrero M, Busch C, Ortega-Garcia J (2019) Biometric
  presentation attack detection: Beyond the visible spectrum. IEEE Transactions
  on Information Forensics and Security 15:1261--1275

\bibitem[{Travieso et~al.(2014)Travieso, Ticay-Rivas, Briceno, del
  Pozo-Ba{\~n}os, and Alonso}]{travieso2014hand}
Travieso CM, Ticay-Rivas JR, Briceno JC, del Pozo-Ba{\~n}os M, Alonso JB (2014)
  Hand shape identification on multirange images. Information Sciences
  275:45--56

\bibitem[{Wang et~al.(2004)Wang, Bovik, Sheikh, and Simoncelli}]{wang2004image}
Wang Z, Bovik AC, Sheikh HR, Simoncelli EP (2004) Image quality assessment:
  from error visibility to structural similarity. IEEE transactions on image
  processing 13(4):600--612

\bibitem[{Wu et~al.(2017)Wu, Yamagishi, Kinnunen, Hanil{\c{c}}i, Sahidullah,
  Sizov, Evans, Todisco, and Delgado}]{wu2017asvspoof}
Wu Z, Yamagishi J, Kinnunen T, Hanil{\c{c}}i C, Sahidullah M, Sizov A, Evans N,
  Todisco M, Delgado H (2017) Asvspoof: the automatic speaker verification
  spoofing and countermeasures challenge. IEEE Journal of Selected Topics in
  Signal Processing 11(4):588--604

\bibitem[{Xia et~al.(2018)Xia, Yuan, Lv, Sun, Xiong, and Shi}]{xia2018novel}
Xia Z, Yuan C, Lv R, Sun X, Xiong NN, Shi YQ (2018) A novel weber local binary
  descriptor for fingerprint liveness detection. IEEE Transactions on Systems,
  Man, and Cybernetics: Systems

\bibitem[{{Xue} et~al.(2014){Xue}, {Zhang}, {Mou}, and {Bovik}}]{6678238}
{Xue} W, {Zhang} L, {Mou} X, {Bovik} AC (2014) Gradient magnitude similarity
  deviation: A highly efficient perceptual image quality index. IEEE
  Transactions on Image Processing 23(2):684--695

\bibitem[{Yoruk et~al.(2006)Yoruk, Konukoglu, Sankur, and
  Darbon}]{yoruk2006shape}
Yoruk E, Konukoglu E, Sankur B, Darbon J (2006) Shape-based hand recognition.
  IEEE transactions on image processing 15(7):1803--1815

\bibitem[{Zhang et~al.(2013)Zhang, Feng, Wang, and Xue}]{zhang2013edge}
Zhang X, Feng X, Wang W, Xue W (2013) Edge strength similarity for image
  quality assessment. IEEE Signal processing letters 20(4):319--322

\bibitem[{{Zhou} et~al.(2019){Zhou}, {Yao}, {Liu}, and {Qiu}}]{8640853}
{Zhou} F, {Yao} R, {Liu} B, {Qiu} G (2019) Visual quality assessment for
  super-resolved images: Database and method. IEEE Transactions on Image
  Processing 28(7):3528--3541

\end{thebibliography}
\end{document}